\documentclass[10pt,twocolumn,letterpaper]{article}

\usepackage{cvpr}              % To produce the CAMERA-READY version
\usepackage{graphicx}
\usepackage{amsmath}
\usepackage{amssymb}
\usepackage{booktabs}

\usepackage[pagebackref,breaklinks,colorlinks]{hyperref}

\usepackage[capitalize]{cleveref}
\crefname{section}{Sec.}{Secs.}
\Crefname{section}{Section}{Sections}
\Crefname{table}{Table}{Tables}
\crefname{table}{Tab.}{Tabs.}

\def\cvprPaperID{***} % *** Enter the CVPR Paper ID here
\def\confName{CVPR}
\def\confYear{2022}

\begin{document}

\title{Network Compression via Central Filter}

\author{ Yuanzhi Duan$^1$, Xiaofang Hu$^{1,3}$\thanks{Corresponding author}\;, Yue Zhou$^{2,3}$, Qiang Liu$^4$, Shukai Duan$^{1,3}$\\
$^1$College of Artificial Intelligence, Southwest University, China\\
$^2$College of Electronics and Information Engineering, Southwest University, China\\
$^3$Brain-inspired Computing \& Intelligent Control of Chongqing Key Lab, China\\
$^4$Harbin Institute of Technology, China\\
{\tt\small swuyzhid@email.swu.edu.cn, \{huxf, duansk\}@swu.edu.cn, zhouyuenju@163.com,}\\
{\tt\small 18b933041@stu.hit.edu.cn}
}
\maketitle

%%%%%%%%% ABSTRACT
\begin{abstract}
Neural network pruning has remarkable performance for reducing the complexity of deep network models.
Recent network pruning methods usually focused on removing unimportant or redundant filters in the network.  
In this paper, by exploring the similarities between feature maps, we propose a novel filter pruning method—Central Filter (CF), which suggests that a filter is approximately equal to a set of other filters after appropriate adjustments. 
Our method is based on the discovery that the average similarity between feature maps changes very little, regardless of the number of input images. 
Based on this finding, we establish similarity graphs on feature maps and calculate the closeness centrality of each node to select the Central Filter.
Moreover, we design a method to directly adjust weights in the next layer corresponding to the Central Filter, effectively minimizing the error caused by pruning.
Through experiments on various benchmark networks and datasets, CF yields state-of-the-art performance.
For example, with ResNet-56, CF reduces approximately 39.7\% of FLOPs by removing 47.1\% of the parameters, with even 0.33\% accuracy improvement on CIFAR-10.
With GoogLeNet, CF reduces approximately 63.2\% of FLOPs by removing 55.6\% of the parameters, with only a small loss of 0.35\% in top-1 accuracy on CIFAR-10.
With ResNet-50, CF reduces approximately 47.9\% of FLOPs by removing 36.9\% of the parameters,  
with only a small loss of 1.07\% in top-1 accuracy on ImageNet.
The codes can be available at \url{https://github.com/8ubpshLR23/Central-Filter}.
\end{abstract}

%%%%%%%%% BODY TEXT
\section{Introduction}\label{introduction}
\begin{figure*}[!t]
\begin{center}
\begin{minipage}[!t]{0.95\linewidth}
\hspace{0.2em}
\centerline{
\begin{subfigure}[]{0.25\linewidth}
	\includegraphics[width=1\linewidth]{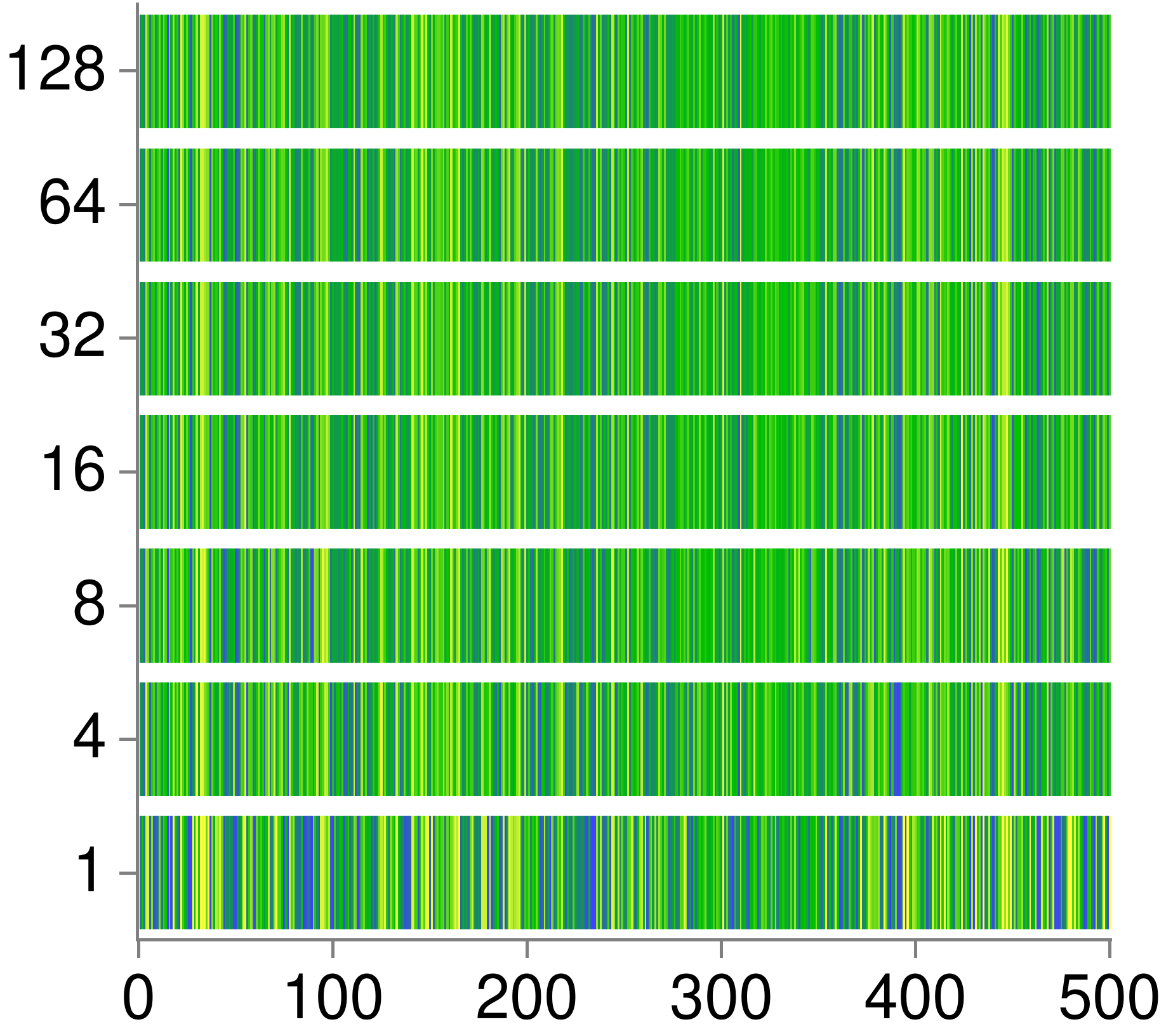}
	\caption{VGGNet-16-1}
\end{subfigure} 
\begin{subfigure}[]{0.25\linewidth}
	\includegraphics[width=1\linewidth]{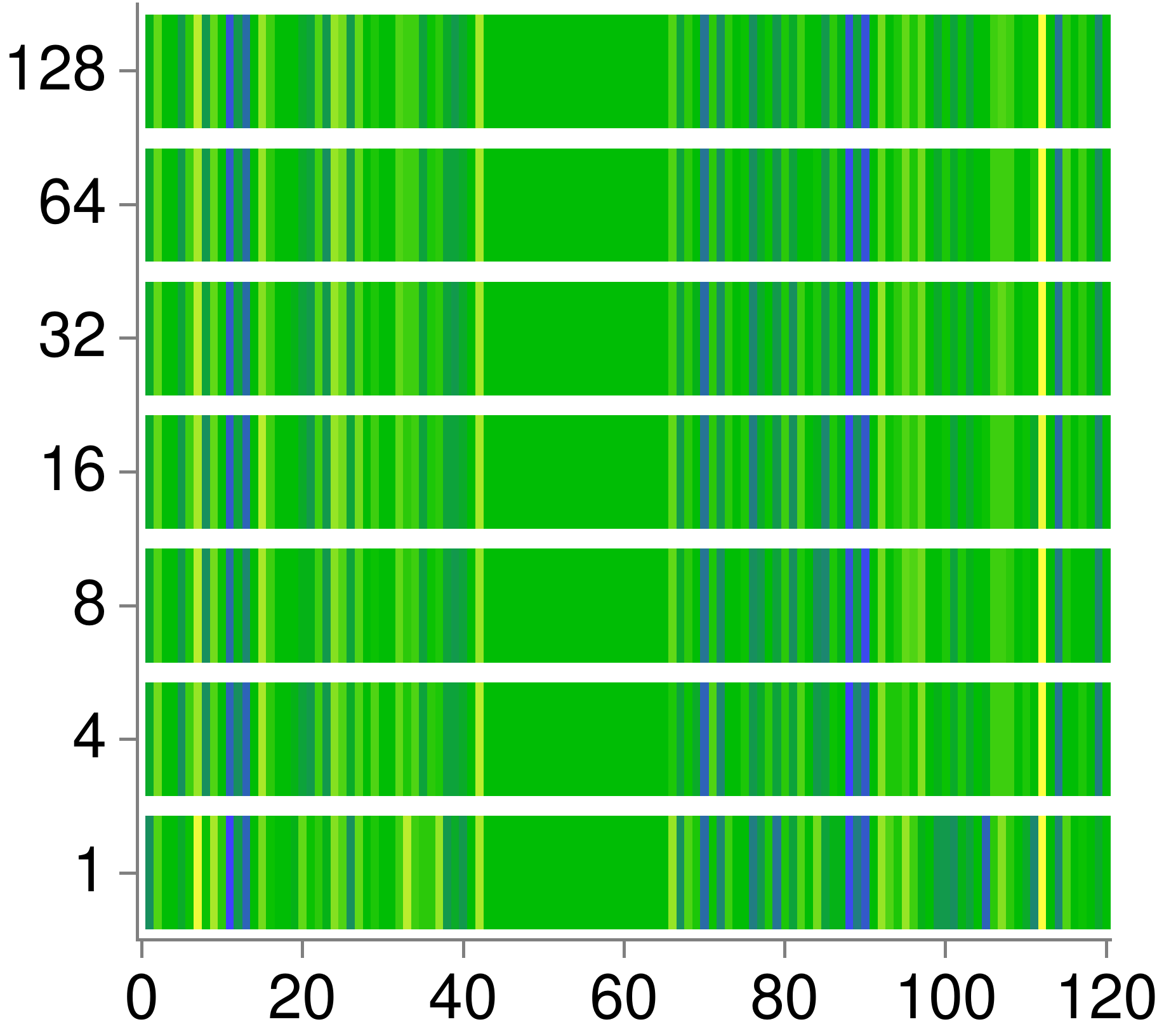}
	\caption{ResNet-56-1}
\end{subfigure}
\begin{subfigure}[]{0.2415\linewidth}
	\includegraphics[width=1\linewidth]{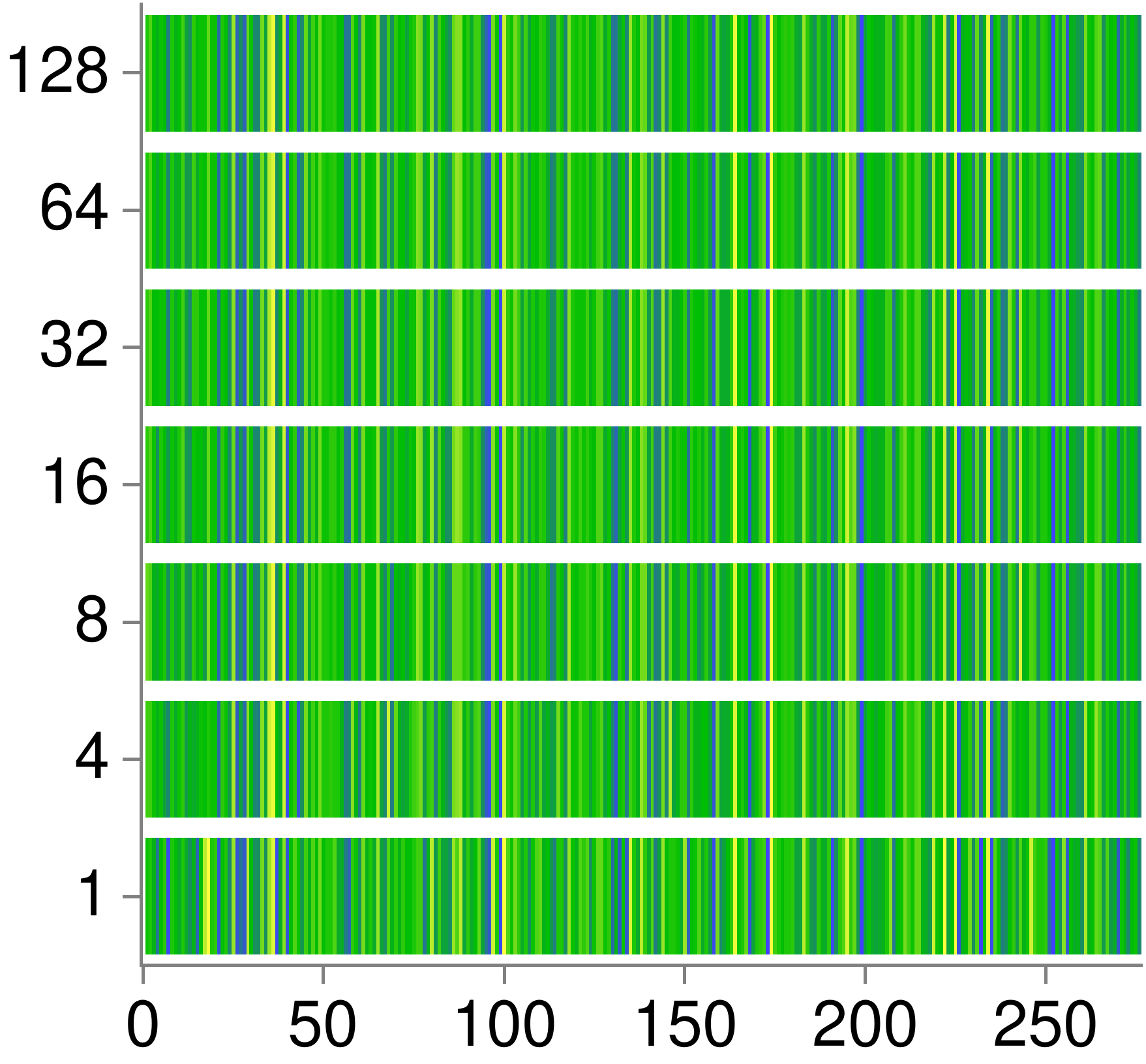}
	\caption{DenseNet-40-1}
\end{subfigure}
\begin{subfigure}[]{0.25\linewidth}
	\includegraphics[width=1\linewidth]{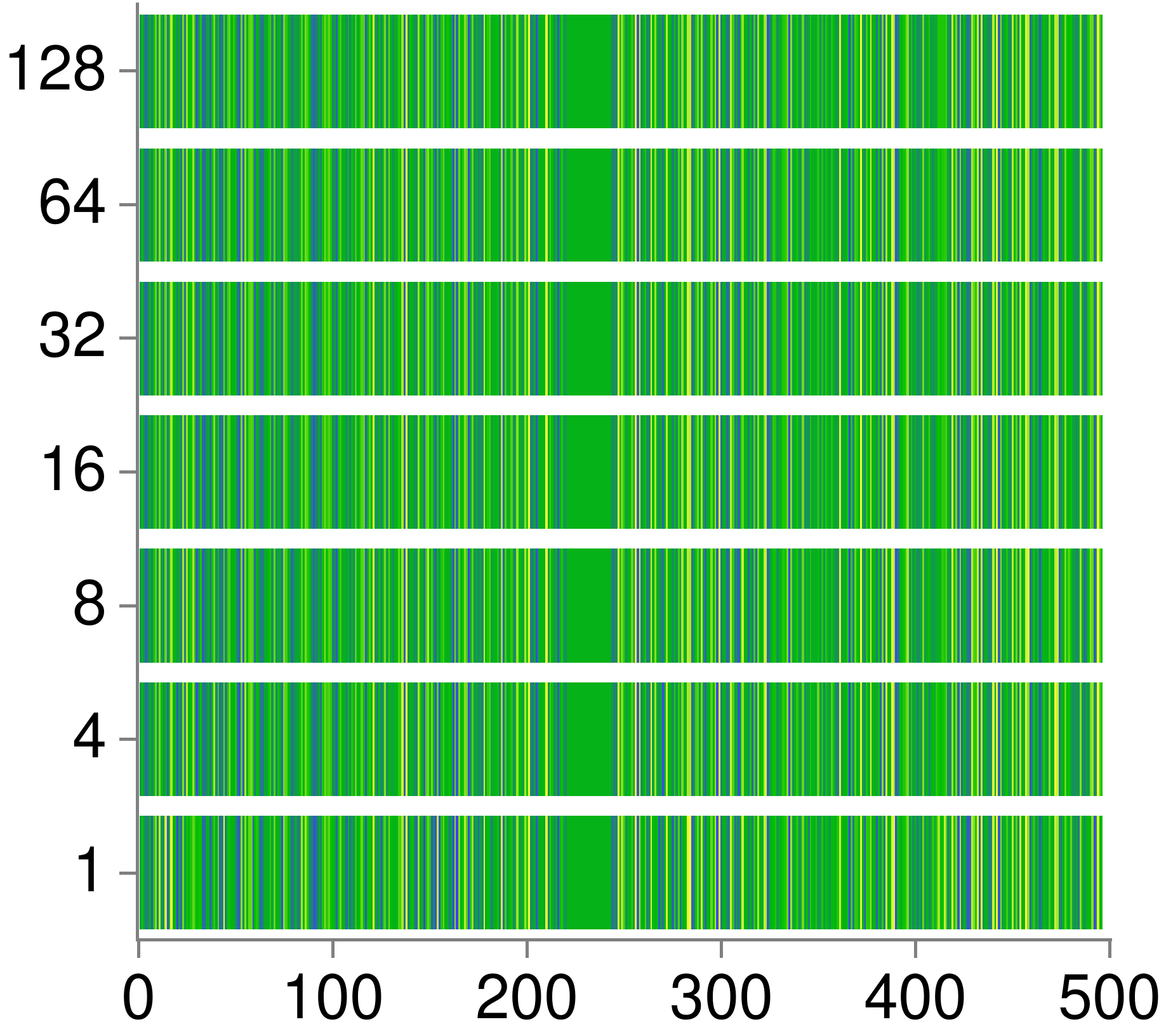}
	\caption{GoogLeNet-1}
\end{subfigure}
}
\end{minipage}
\end{center}
\vspace{-0.8em}
\caption{\label{anls_vgg}Average similarity statistics of output feature maps of the first layer from different models on CIFAR-10. For each subfigure, the $x$-axis represents the different pairs of features, and the $y$-axis is the number of training images. Different colors denote different similar values. As we can see, there are many feature map pairs with almost constant values, regardless of the number of input images. Hence, even a small set of images can effectively estimate the average similarity of each feature pair in different models.}
\vspace{-1.2em}
\end{figure*}

\begin{figure}[!t]
\begin{center}
\begin{minipage}[!t]{1.0\linewidth}
\hspace{0.5em}
\centerline{
\begin{subfigure}[]{1.1\linewidth}
	\includegraphics[width=1\linewidth]{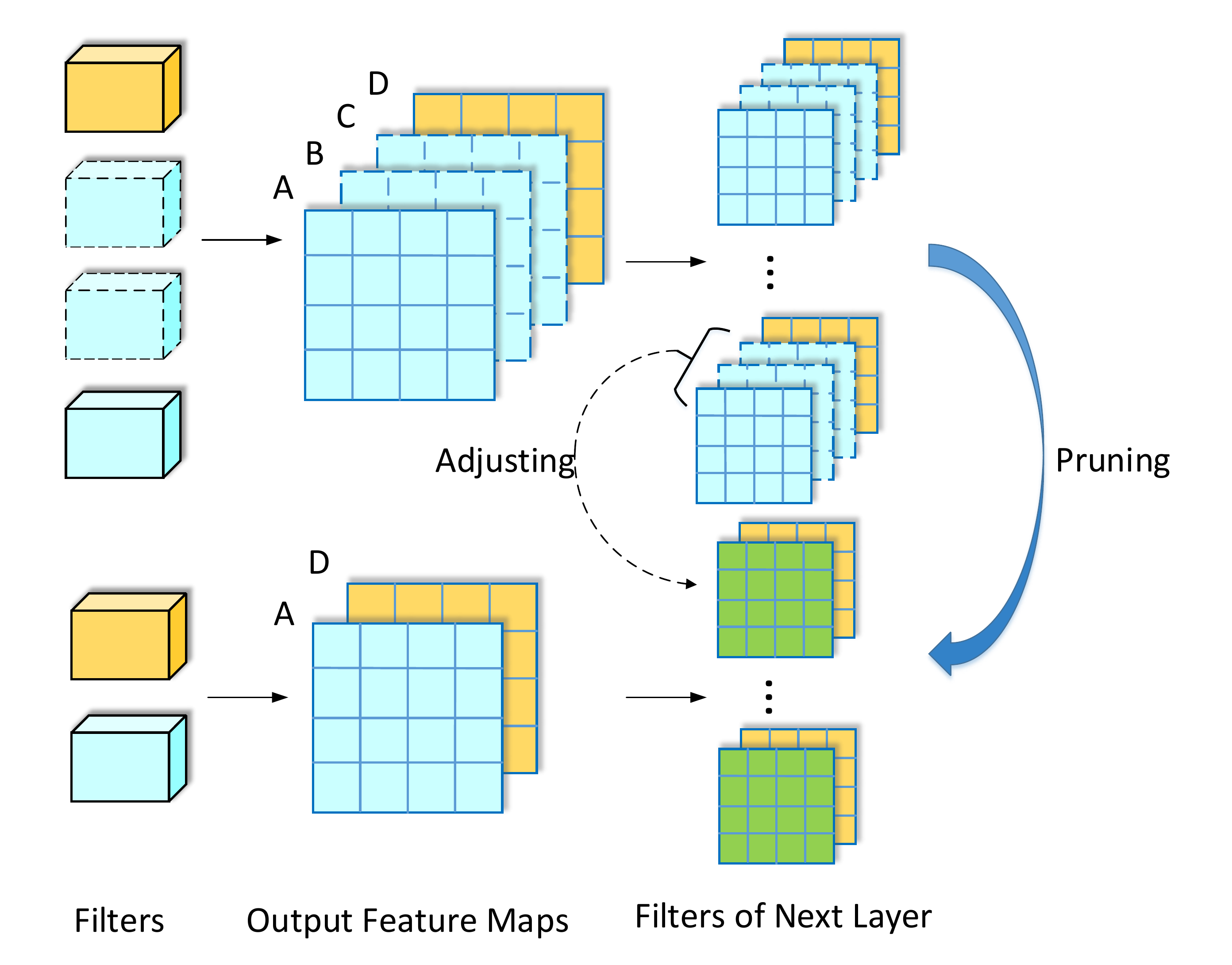}
	\caption{}
\end{subfigure} 
}
\end{minipage}
\begin{minipage}[!t]{1.0\linewidth}
\hspace{0.3em}
\centerline{
\begin{subfigure}[]{1\linewidth}
	\includegraphics[width=0.8\linewidth]{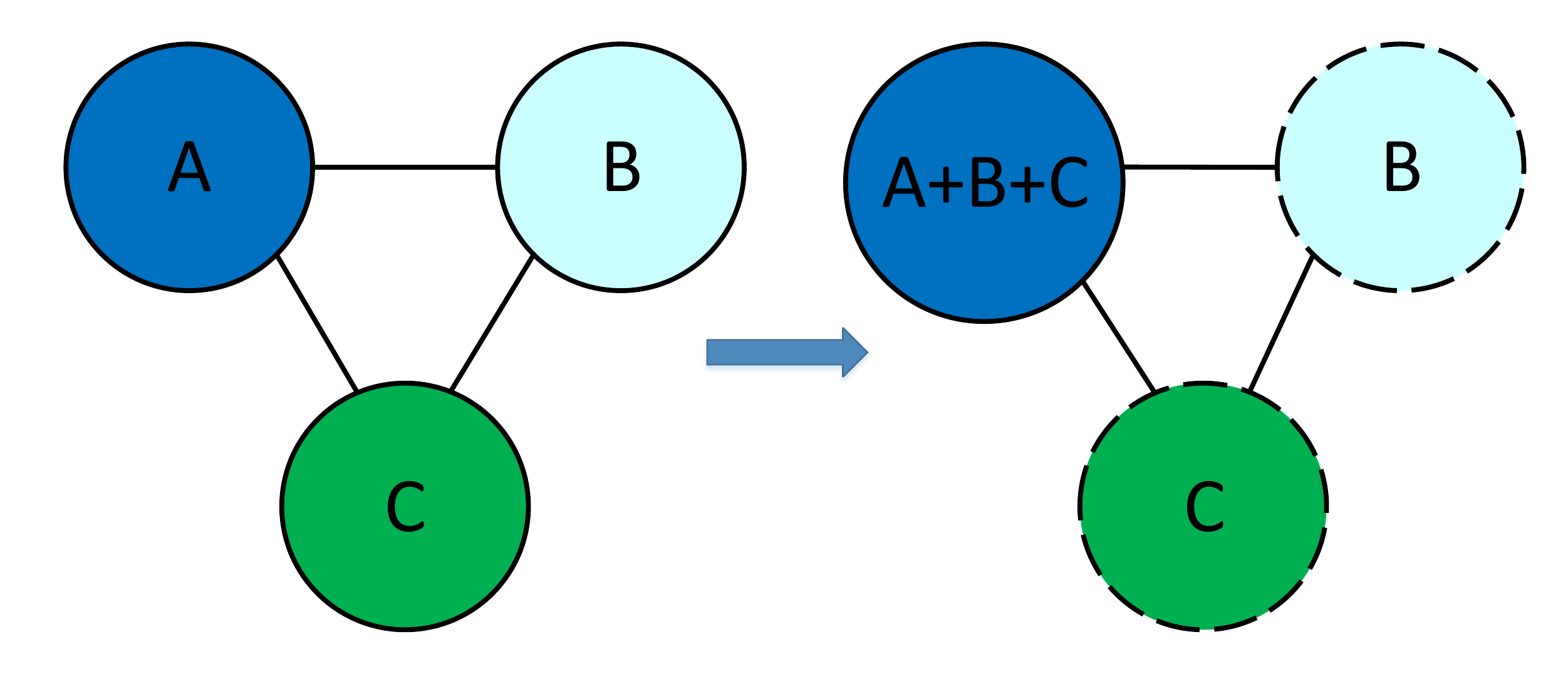}
	\caption{}
\end{subfigure} 
}
\end{minipage}
\end{center}
\vspace{-0.8em}
\caption{\label{central filter process}In (a), the feature maps A, B, and C are similar to each other. Assuming that A is reserved, B and C are pruned. We prune the corresponding filters in this layer and the corresponding kernels in the filters of the next layer. Then we adjust the kernel corresponding to A to A+B+C, making A play the role of A+B+C, as shown in (b). In other words, if A, B, and C are the same, then there will be no loss of accuracy when B and C are pruned. Finally, we use a smaller filter to implement the function of the original filters and call it a central filter.
}
\vspace{-1.2em}
\end{figure}

Recent developments on Convolutional neural networks (CNNs) have achieved great performance in computer vision tasks such as  image classification \cite{He2016DeepRL,Huang2017DenselyCC}, object detection \cite{Ren2015FasterRT,Girshick2014RichFH}, semantic segmentation \cite{Long2015FullyCN,Chen2018DeepLabSI}, video analysis \cite{Girdhar2019DistInitLV,Lin2019BMNBN}, etc.
However, state-of-the-art CNNs have higher requirements for computational resources and memory footprint, which greatly limits their application in edge devices such as IoT devices.
To address this, one idea is to compress the existing network, including tensor decomposition \cite{Phan2020StableLT}, network pruning \cite{Lin2020HRankFP}, parameter quantification \cite{Liu2020ReActNetTP}, etc. Another idea is to build a new small network, including knowledge distillation \cite{Hinton2015DistillingTK} and compact network design \cite{Howard2017MobileNetsEC}. 
Among these network compression strategies, network pruning has significant performance and is suitable for various applications.
In terms of the granularity of network pruning, it can be divided into unstructured pruning \cite{CarreiraPerpin2018LearningCompressionAF,Han2015LearningBW} and structured pruning \cite{Lin2020HRankFP,Luo2017ThiNetAF}.   
Some early methods were based on unstructured pruning, and the produced kernels were sparse. That is, there are many matrices with zero elements in the middle. 
Unless the underlying hardware and computing libraries support them better, it is difficult to achieve substantial performance gains in the pruned models.
Besides, the sparse matrix cannot use the existing mature BLAS library to obtain additional performance benefits.
Therefore, a lot of research in recent years has focused on structured pruning, especially filter pruning. 
The basic idea of filter pruning is to prune the unimportant filters while minimizing the loss of accuracy.
So, the challenge is how to measure the importance of filters. 
For example, Li \emph{et al}. \cite{Li2017PruningFF} assumed that a filter should be pruned first if it has a smaller $l_1$ norm.
Liu \emph{et al}. \cite{Liu2017LearningEC} established a link between the importance of the filter and the scaling factor of the BN (Batch Normalization) layer. Filters with a smaller scaling factor value are unimportant and will be pruned first.
In \cite{Lin2020HRankFP}, the High Rank (HRank) of the feature map is used to measure the importance of the filters in each layer.
These methods are based on different experiences or assumptions and succeed in filter pruning.
Another idea is to focus on the redundancy of filters or feature maps while pruning.
Deep neural networks have a lot of redundancy \cite{Denil2013PredictingPI}. So empirically, some redundant filters can be removed for network compression.  
A popular solution is to consider the similarity of filters or feature maps.
Singh \emph{et al}. \cite{Singh2020LeveragingFC} searched for strongly correlated filter pairs from each layer and made them highly correlated to reduce accuracy loss, and finally discard one.
Wang  \emph{et al}. \cite{Wang2021ModelPB}, proposed Quantified Similarity of Feature Maps (QSFM) to find the redundant information in the three-dimensional tensors.
Most similarity-based designs achieve compression and acceleration of neural networks, but with a limitation in the assumption that redundancy is useless.
Our work is based on the similarity of the feature maps, which belong to the filter pruning.
Many of the existing methods are based more or less on assumptions or experiences that lack theoretical proof.
In this paper, we propose a novel approach to prune redundant filters. 
Firstly, we explored the stability of the similarity between feature maps in each layer. 
As shown in Fig.\;\ref{anls_vgg}, we calculated the average similarity of output feature maps of different models on CIFAR-10 with different numbers of input images.
After multiple sets of experiments, we found that the similarity changes between different feature maps are few and can be ignored. 
The average rank of feature maps has been explored in Hrank \cite{Lin2020HRankFP}, showing that the rank of each feature map is always the same in different models. 
This proves the stability of the properties of the feature map from another perspective.
Then we propose a novel theory, named Central Filter(CF), to selectively prune filters that generate redundant feature maps. 
In a set of similar feature maps, we select one feature to cover the functions of other features  by directly adjusting the corresponding filter weights.
To support this, mathematical proof is provided.
As we all know, there is a lot of redundancy in many models \cite{Denil2013PredictingPI}, so removing the redundancy is a common solution for network compression. 
However, simply removing redundancy may have inadequacies. 
For example, there may be information supplementation between similar feature maps \cite{Singh2020LeveragingFC}, which may result in loss of accuracy when discarding some of them.
The key point of our approach is to create the central filter by analyzing the relationships between feature maps and by directly adjusting the filter weights, which reduces the potential effect of pruning redundant filters, as will be discussed in Sec.\;\ref{the_proposed_method}.
Moreover, a critical issue is how to determine which filters are central filters.
As shown in Fig.\;\ref{central filter process}, we present an approach based on the closeness centrality of graph theory to solve it, and will be discussed in Sec.\;\ref{Filter Selection}. 
Finally, we iteratively prune similar filters layer by layer and directly adjust the weight of filters in the next layer, and then through fine-tuning to recover the accuracy of the pruned model.
To summarize, our main contributions are as follows:
\begin{itemize}
  \item Through a large number of experiments in different models, We empirically demonstrate that the average similarity between the output feature maps is stable, regardless of the number of input features. Inspired by this, we propose a method to use this similar information, named Central Filter. 
  \item We mathematically prove the rationality of the proposed method. More, we present an approach based on the closeness centrality of graph theory to determine which filters should be pruned and which weights should be adjusted.
  \item Extensive experiments on CIFAR-10 \cite{Krizhevsky2009LearningML} and ImageNet \cite{Krizhevsky2012ImageNetCW}, using VGGNet \cite{Simonyan2015VeryDC}, GoogLeNet \cite{Szegedy2015GoingDW}, DenseNet \cite{Huang2017DenselyCC}, and ResNet \cite{He2016DeepRL}, demonstrate the effectiveness of the proposed Central Filter method.
\end{itemize}
%
%%%%%%%%%%%%%%%%%%%%%%%%%%%%%%%%%%%%%%%%%%%%%%%%%%%%%%%%%%%%%%%%%%%%%%%%%%%%%
%%%%%%%%%%%%%%%%%%%%%%%%%%%%%%%%%%%%%%%%%%%%%%%%%%%%%%%%%%%%%%%%%%%%%%%%%%%%%
%%%%%%%%%%%%%%%%%%%%%%%%%%%%%%%%%%%%%%%%%%%%%%%%%%%%%%%%%%%%%%%%%%%%%%%%%%%%%
\begin{figure*}[th]
\begin{center}
\begin{minipage}[t]{0.8\linewidth}
\centerline{
\begin{subfigure}[]{0.25\linewidth}
	\includegraphics[width=1\linewidth]{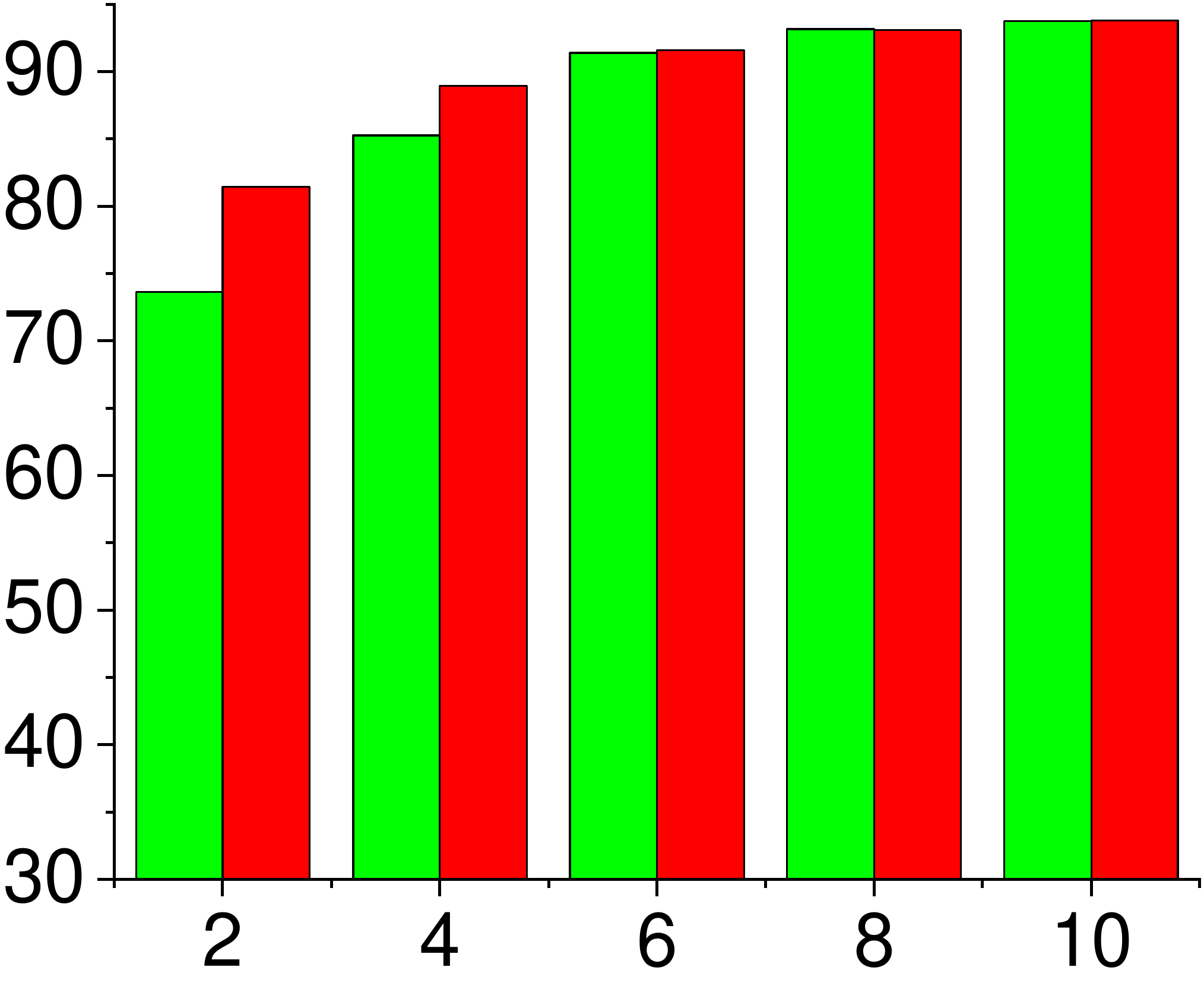}
	\caption{VGGNet-16 $\lambda=0.3$.}
\end{subfigure} 
\begin{subfigure}[]{0.25\linewidth}
	\includegraphics[width=1\linewidth]{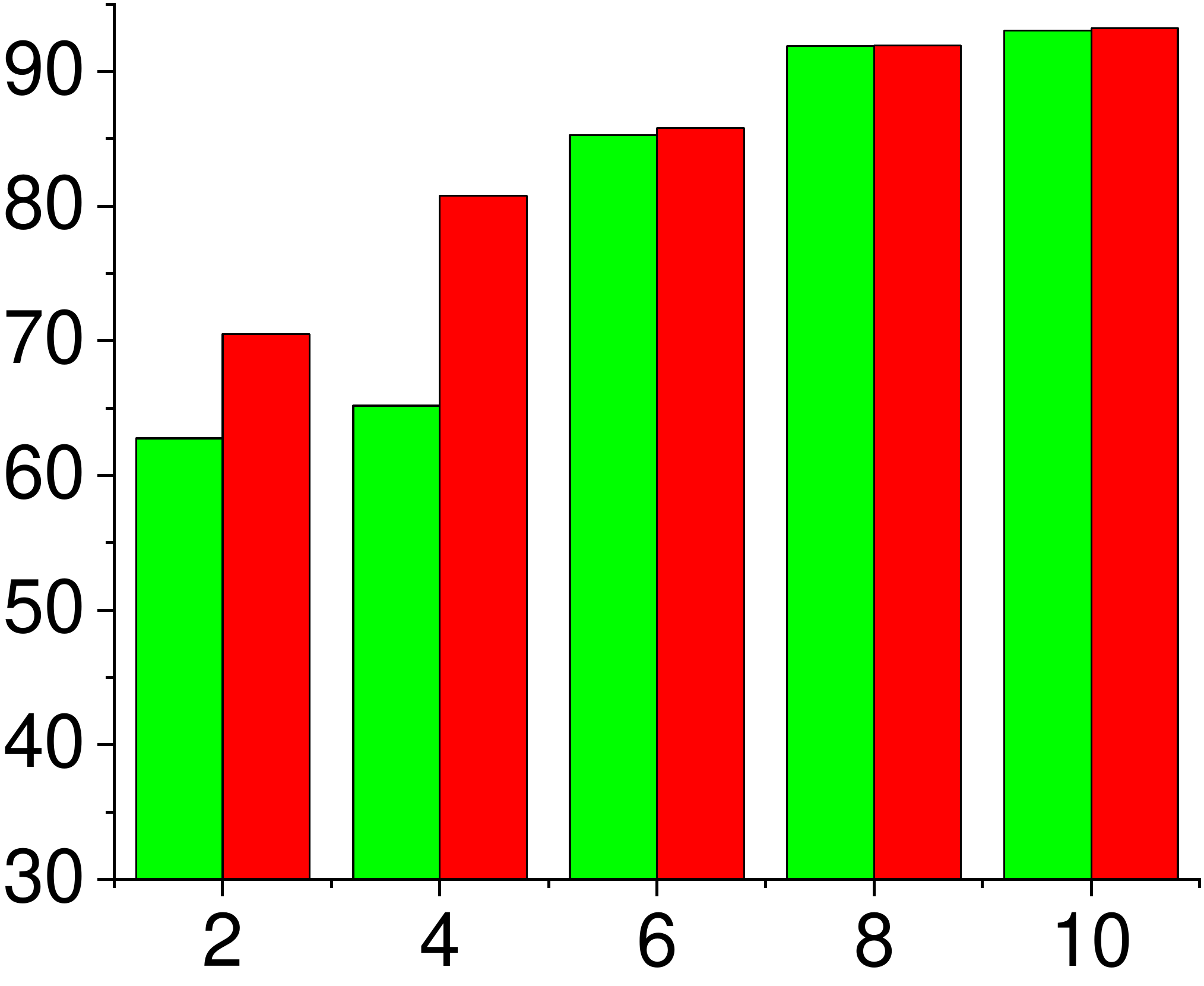}
	\caption{VGGNet-16 $\lambda=0.5$.}
\end{subfigure} 
\begin{subfigure}[]{0.25\linewidth}
	\includegraphics[width=1\linewidth]{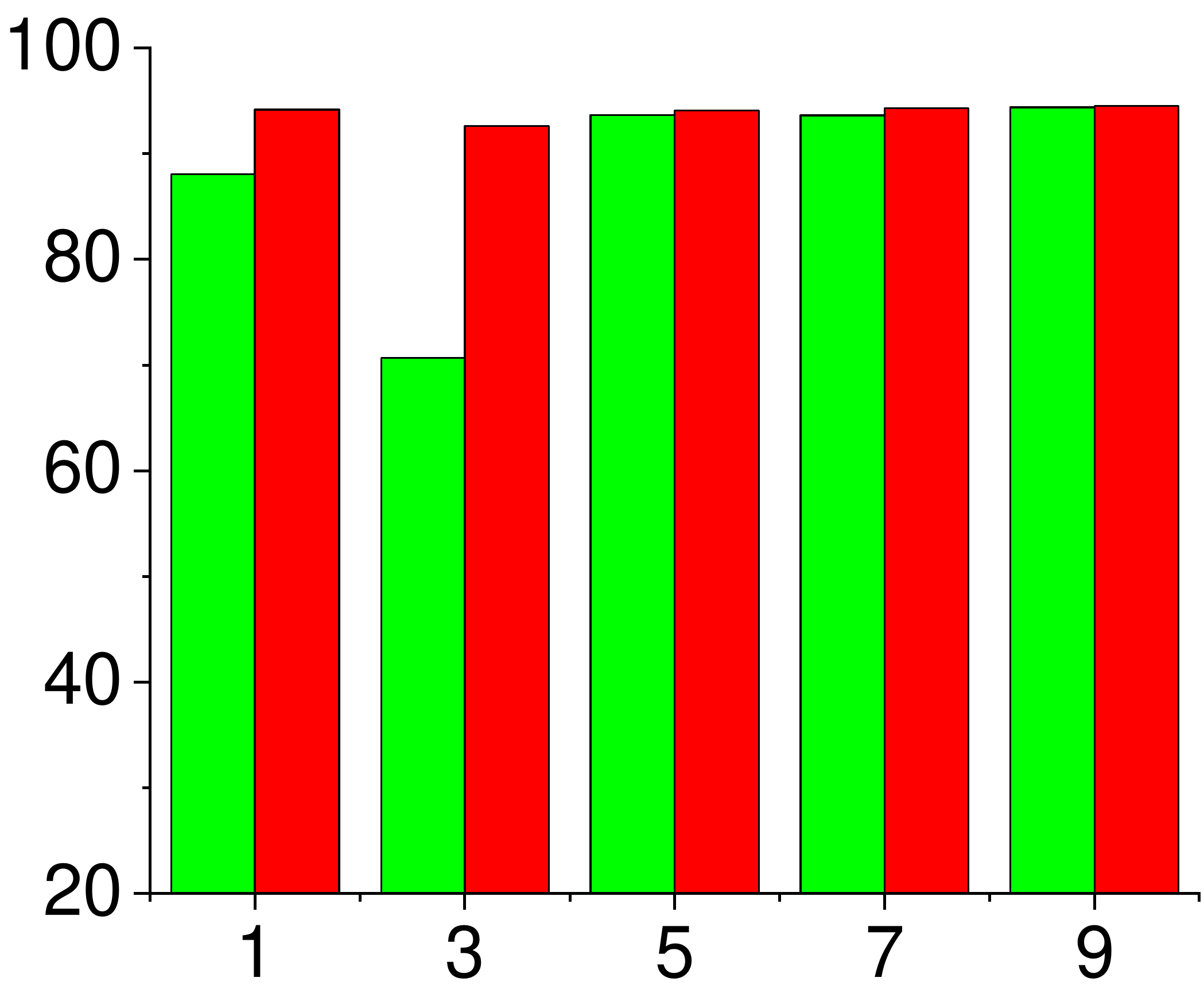}
	\caption{GoogLeNet $\lambda=0.3$.}
\end{subfigure} 
\begin{subfigure}[]{0.25\linewidth}
	\includegraphics[width=1\linewidth]{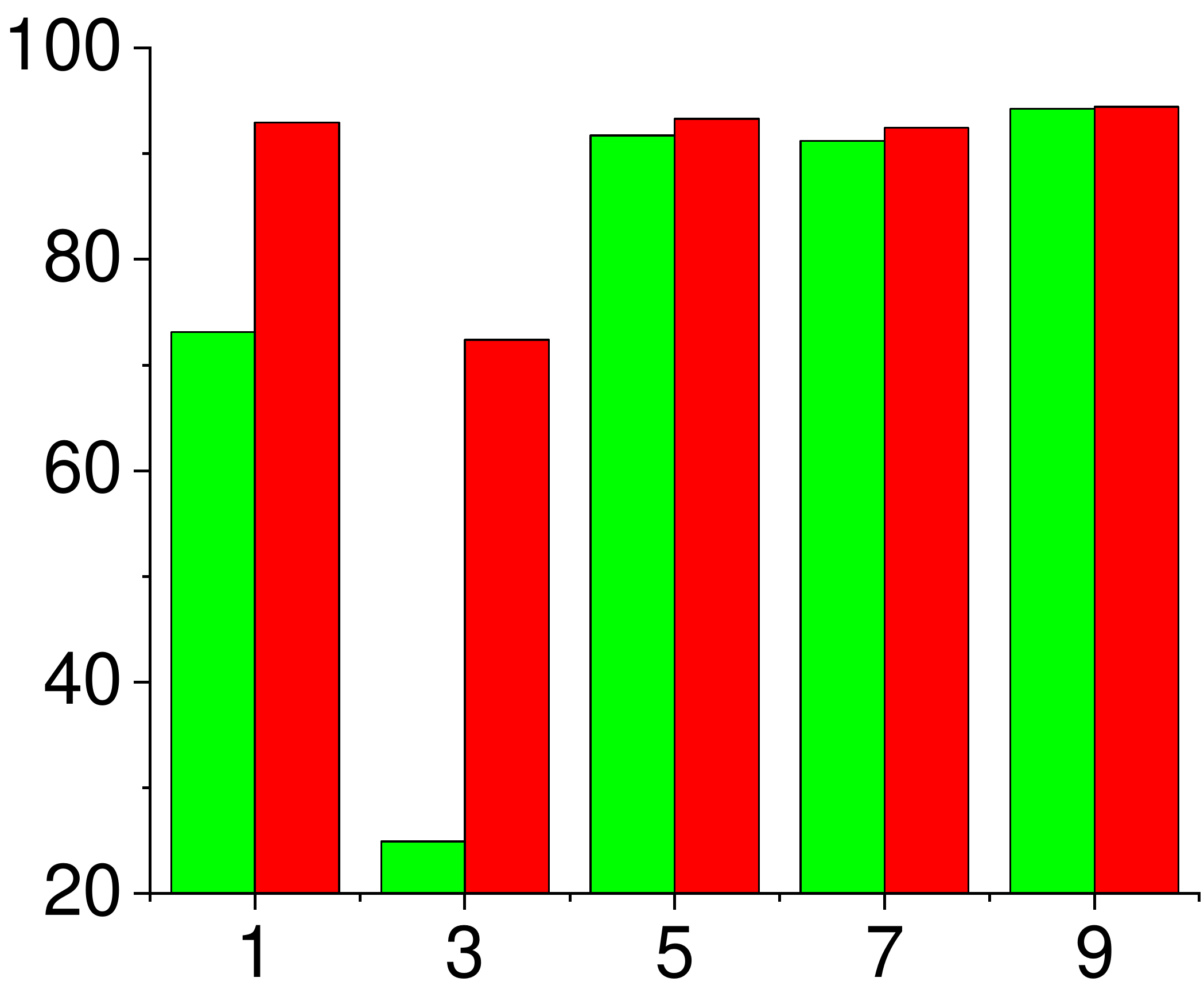}
	\caption{GoogLeNet $\lambda=0.5$.}
\end{subfigure} 
}
\end{minipage}

\begin{minipage}[t]{0.8\linewidth}
\centerline{
\begin{subfigure}[]{0.25\linewidth}
	\includegraphics[width=1\linewidth]{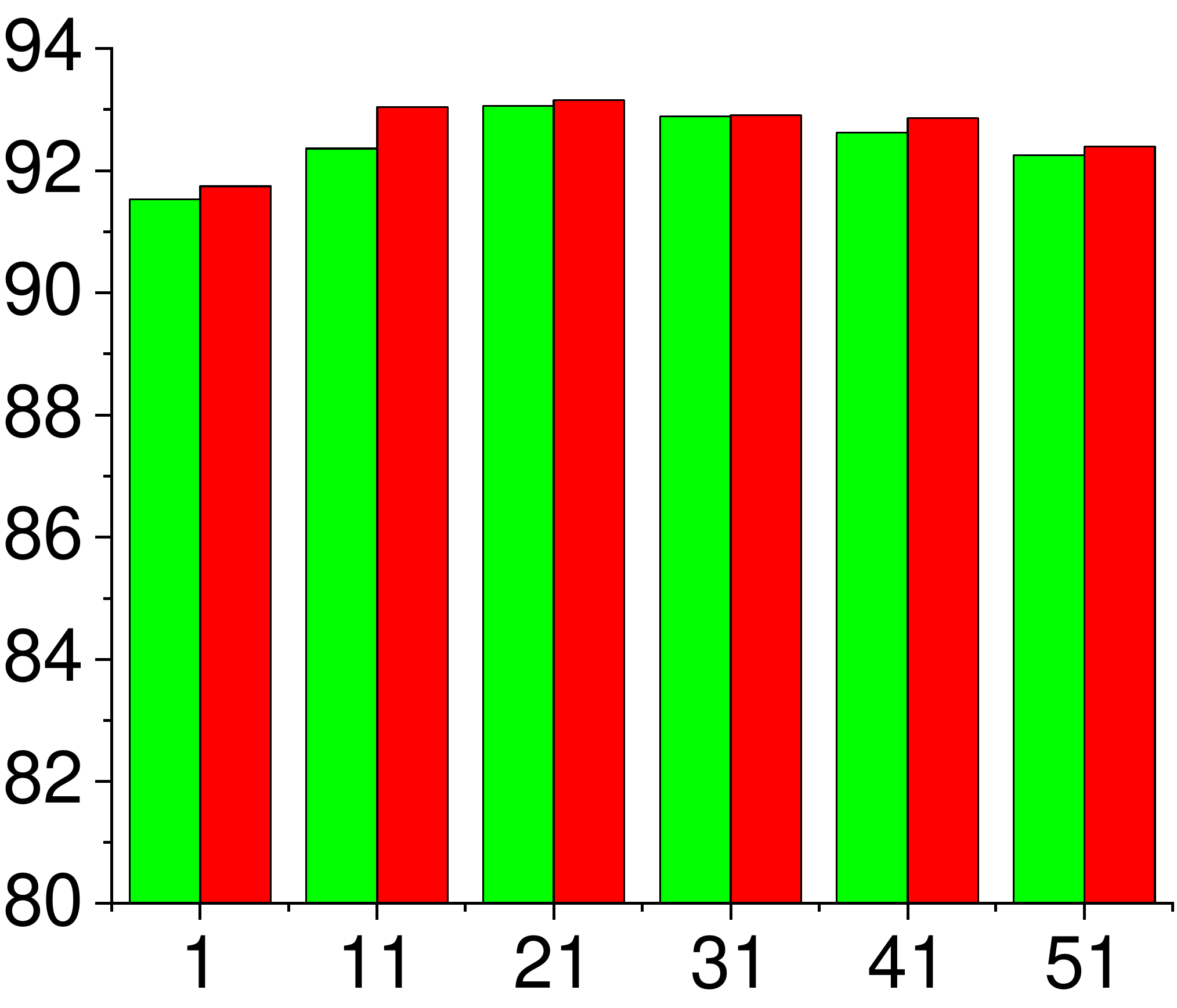}
	\caption{ResNet-56 $\lambda=0.3$.}
\end{subfigure} 
\begin{subfigure}[]{0.25\linewidth}
	\includegraphics[width=1\linewidth]{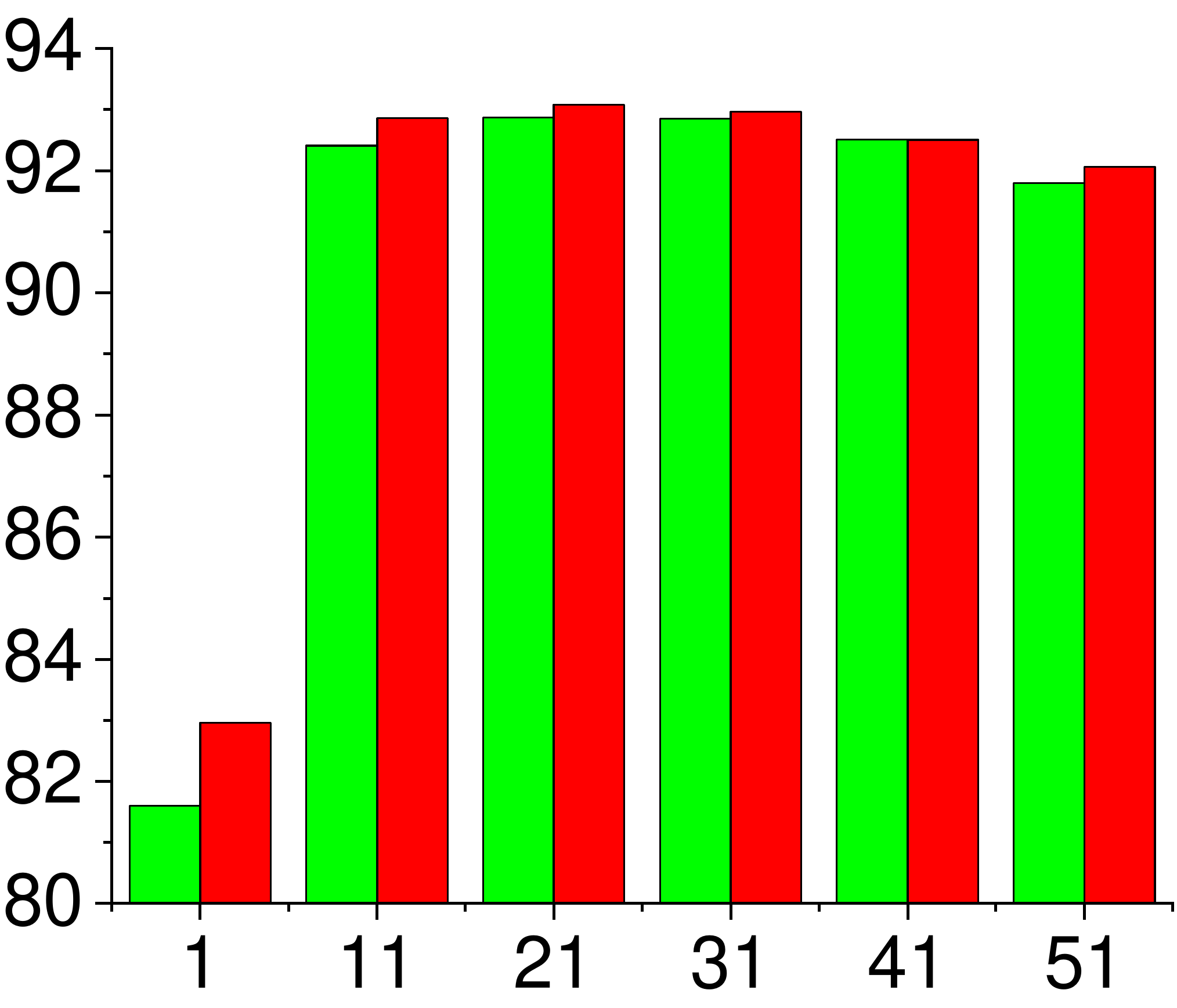}
	\caption{ResNet-56 $\lambda=0.5$.}
\end{subfigure} 
\begin{subfigure}[]{0.25\linewidth}
	\includegraphics[width=1\linewidth]{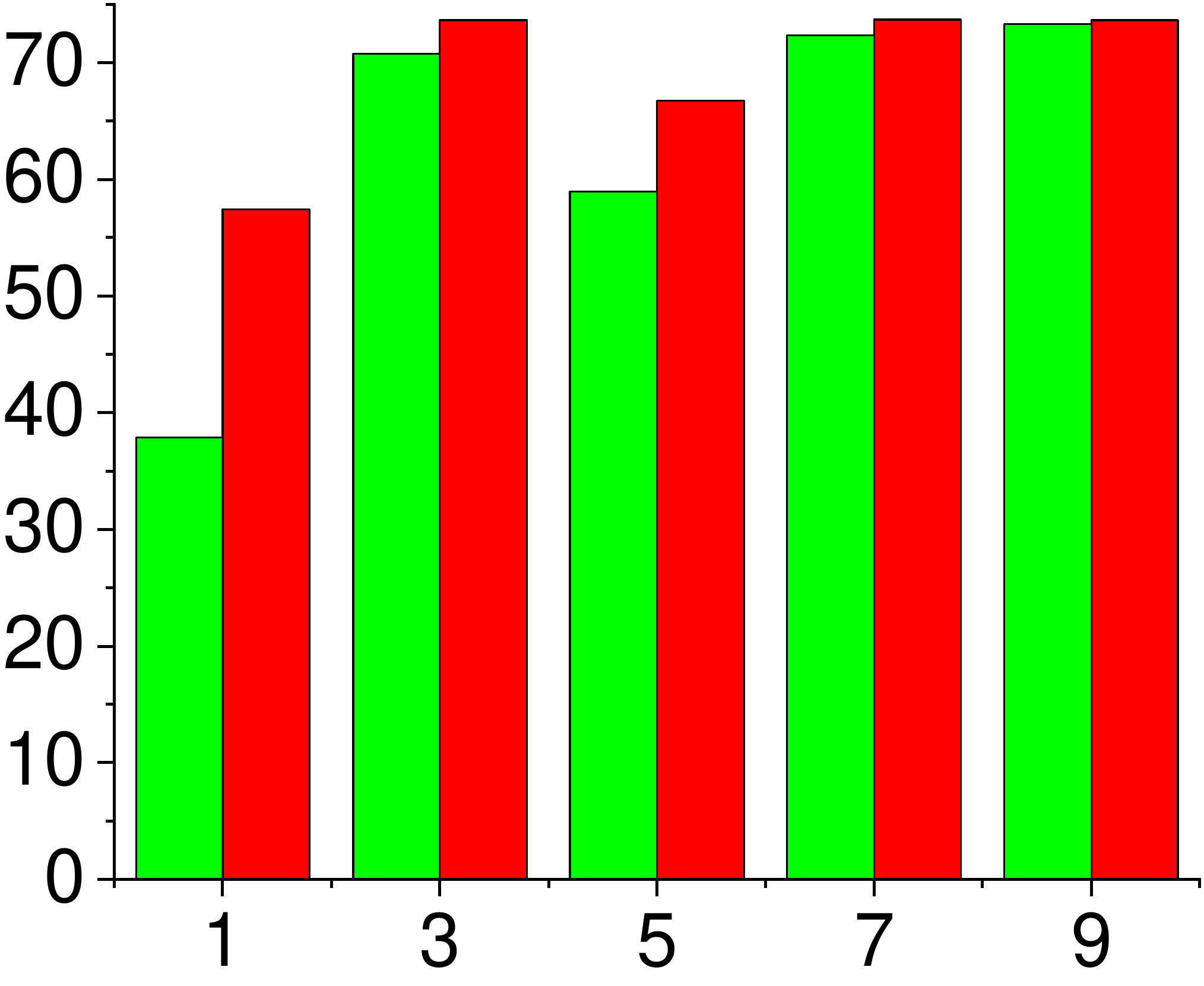}
	\caption{ResNet-50 $\lambda=0.3$.}
\end{subfigure} 
\begin{subfigure}[]{0.25\linewidth}
	\includegraphics[width=1\linewidth]{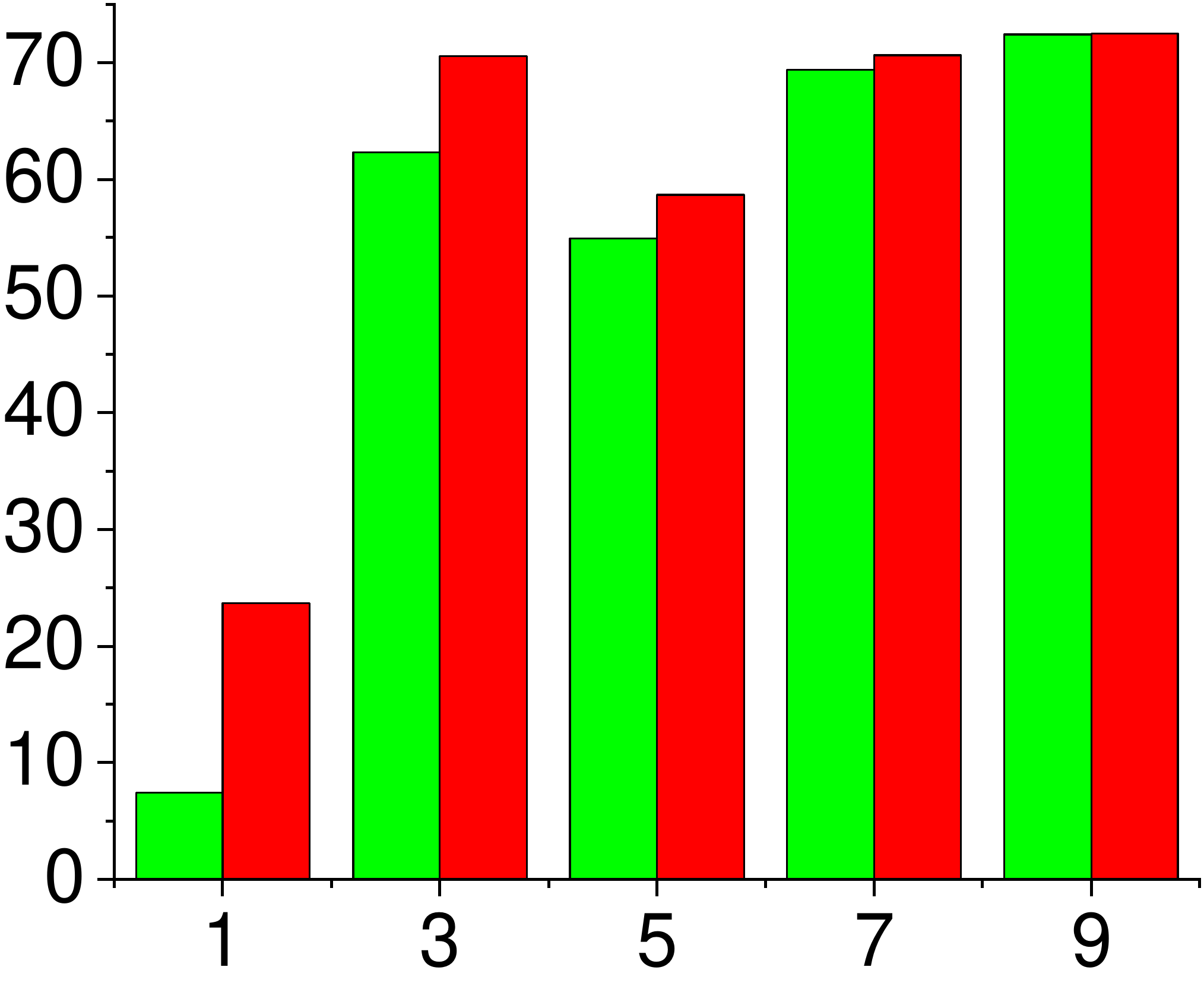}
	\caption{ResNet-50 $\lambda=0.5$.}
\end{subfigure} 
}
\end{minipage}

\end{center}
\vspace{-0.8em}
\caption{\label{filter adjust}The comparison of the accuracy before and after the adjustment of Central Filter from different convolutional layers and CNN models on CIFAR-10 or ImageNet. For each subfigure, the $x$-axis represents the $i$-th layer and the $y$-axis is the accuracy, and $\lambda$ means compression rate. Furthermore, columns with green denote the accuracy of before adjustment, and red is the after. As can be seen, the accuracy has been improved after adjustment on each layer in different architectures.}
\vspace{-1.2em}
\end{figure*}
%%%%%%%%%%%%%%%%%%%%%%%%%%%%%%%%%%%%%%%%%%%%%%%%%%%%%%%%%%%%%%%%%%%%%%%%%%%%%
%-------------------------------------------------------------------------
\section{Related Work}\label{Related Work}
\textbf{Filter Pruning Based on Importance}.
Filter pruning approaches lie in the evaluation of the importance of filters.
Many classic methods measure importance by studying and analyzing the intrinsic properties of filters and feature maps. 
For example, Li \emph{et al}. \cite{Li2017PruningFF} established a connection between the importance of the filter and the  $l_1$ norm, which assumes that a filter is unimportant if it has a smaller $l_1$ norm.
He \emph{et al}. \cite{He2018PruningFV} points out the limitations of the $l_1$ norm-based methods, and the geometric median is used to measure the importance of filters, which the filters closest to it are pruned first.
In \cite{Lin2020HRankFP}, the rank of the output feature map is used to measure the importance of filters in each layer, which demonstrates that the feature maps with lower rank are unimportant.
Beyond ranking the filters in each layer, Chin \emph{et al}. \cite{Chin2020TowardsEM} proposed a method to calculate the global ranking of ﬁlters across different layers.
Based on the widespread use of the BN (Batch Normalization) layer, Liu \emph{et al}. \cite{Liu2017LearningEC} added a channel-wise scaling factor to the BN layer and added an $l_1$ regularizer to make it sparse, then pruning the filters with a small scaling factor.
In addition, another direction considers the subsequent impact after pruning, such as the loss of accuracy, or the changes of output feature maps, etc. 
In \cite{Luo2017ThiNetAF,He2017ChannelPF}, pruning filters by computing the statistical information of the next layer, and aims to minimize the feature reconstruction error where filters with a smaller error are discarded first.
\textbf{Filter Pruning Based on Redundancy}.
Different from the previous methods based on importance, which pruning filters by designing a filter evaluation function. Redundancy-based methods concern which filters are redundant and how to remove them.
\cite{Singh2020LeveragingFC} proposed an approach to identify the pairs of strongly correlated filters and discard one from each such pair. The pruned filter is considered redundant because the performance capabilities of the two filters in the pair are the same.
Wang \emph{et al}. \cite{Wang2021ModelPB} proposed a novel theory, namely Quantified Similarity of Feature Maps (QSFM), to get redundant information in the high-dimensional tensors.
In \cite{Shao2021ADC}, cosine similarity is used to measure the similarity of feature maps or filters, and filters with high similarity are pruned.
Wang \emph{et al}. \cite{Wang2021ConvolutionalNN} discovered that the layers with more structural redundancy are more suitable to be pruned, and graph theory is used to measure the redundancy for each layer.  
Though redundancy-based methods succeed in the compression and acceleration of different models, it is still inefficient at high compression rates because of missing theoretical proof and guidance.
Our method also belongs to filter pruning and is based on redundancy. 
Compared with the previous redundancy-based methods, we amply use the properties of similar output feature maps and directly adjust the relevant weights, achieving higher accuracy at the same compression rate.
%
%-------------------------------------------------------------------------
\section{The Proposed Method}\label{the_proposed_method}
\subsection{Notations}\label{Notations}
For a CNN model, which has $N$ layers. We let $\mathcal{L}^i$ represent the $i$-th layer, and $\mathcal{F}_{\mathcal{L}^i}=\{f_1^i,f_2^i,...,f_{n^i}^i\}\in \mathbb{R}^{n^{i-1}\times n^{i} \times h^i \times w^i}$ denotes the set of all filters of the $i$-th layer, where  $n^{i}$ denote the numbers of filters, and $h^i , w^i$ represent the size of the kernel of the $i$-th layer.
We define the $j$-th filter $f_j^i=\{k_1^{i,j},k_2^{i,j},...,k_{n^{i-1}}^{i,j}\}\in \mathbb{R}^{n^{i-1}\times h^i \times w^i}$, where the $t$-th kernel $k_t^{i,j}\in \mathbb{R}^{h^i \times w^i}$.
Let $\mathcal{O}_{\mathcal{L}^i} = \{o_1^i,o_2^i,...,o_{n^i}^i\} \in \mathbb{R}^{c \times n^{i} \times h^i \times w^i}$ be the set of  output features in the $i$-th layer, where $o_{j}^i$ is the $j$-th feature involved the $j$-th filter.
$c$ is the size of input feature maps.
We define $\mathcal{S}(o_{x}^i,o_{y}^i),(0 \leq x,y \leq n^i)$ as the measure of similarity between output features.
%
%-------------------------------------------------------------------------
\subsection{Feature Similarity Measure}\label{the_proposed_approach}
In this paper, we adopt Pearson correlation coefficient to measure the similarity between output feature maps. 
The Pearson correlation coefficient is used to measure the degree of correlation between two variables $X$ and $Y$, and is defined as the covariance of two variables divided by the product of their standard deviations.  
\begin{equation}\label{Pearson correlation coefficient}
\begin{split}
\rho_{X,Y}= \frac{cov(X,Y)}{\sigma_{X} \sigma_{Y}} = \frac{E[(X-\mu_{X})(Y-\mu_{Y})]}{\sigma_{X} \sigma_{Y}} 
\end{split}
\end{equation}
Here $\mu_{X},\mu_{Y}$ represent expected values, and $\sigma_{X}, \sigma_{Y}$ are standard deviations respectively.
The Pearson correlation coefficient varies from $-1$ to $1$. The value of the coefficient is $1$ means that $X$ and $Y$ can be well described by the straight-line equation. All the data points fall well on a straight line, and $Y$ increases with the increase of $X$. The value of the coefficient $-1$ means that all data points fall on a straight line, and $Y$ decreases with $X$ increases. The value of the coefficient is $0$ means that there is no linear relationship between the two variables. So, the similarity $\mathcal{S}(o_{x}^i,o_{y}^i)$ can  be formulated as:
\begin{equation}\label{similarity}
\begin{split}
\mathcal{S}(o_{x}^i,o_{y}^i) = \rho_{One(o_{x}^i),One(o_{y}^i)}
\end{split}
\end{equation}
where $One(\cdot)$ reshape $o_{j}^i$ into one dimension.
For features similarity measure, the value of the coefficient is more near $1$ means two features are more similar. When the value is near $0$, the two features are not similar.
Through a large number of experiments with different models on CIFAR-10, we have observed that the similarity between the output feature maps is stable. 
As demonstrated in Fig.\;\ref{anls_vgg}, we can effectively estimate the expectation of the similarity between individual feature maps in different models with a small number of input images.
Then, using this similarity information, we design an approach called Central Filter to determine which filters to discard
%
%-------------------------------------------------------------------------
\subsection{Central Filter}\label{CentralFilter}
For the $i$-th layer of the CNN model, the output features can be formulated as:
\begin{equation}\label{output features}
\begin{split}
\mathcal{O}_{\mathcal{L}^{i}} = \mathcal{F}_{\mathcal{L}^i} \cdot \mathcal{O}_{\mathcal{L}^{i-1}}
 = \{f_1^i,f_2^i,...,f_{n^i}^i\} \cdot \mathcal{O}_{\mathcal{L}^{i-1}}
\end{split}
\end{equation}
For simplicity, functions such as Relu, Batch normalization are omitted here.
For the $j$-th output feature $o_j^i$ :
\begin{equation}\label{output feature}
\begin{split}
o_j^{i} &=  f_j^i \cdot \mathcal{O}_{\mathcal{L}^{i-1}} \\
&= \{k_1^{i,j},k_2^{i,j},...,k_{n^{i-1}}^{i,j}\} \cdot \{o_1^{i-1},o_2^{i-1},...,o_{n^{i-1}}^{i-1}\} \\
&= \sum\limits_{r=1}^{n^{i-1}}k_r^{i,j} * o_r^{i-1} \\
&= k_1^{i,j} * o_1^{i-1} + k_2^{i,j} * o_2^{i-1}+...+ k_{n^{i-1}}^{i,j} * o_{n^{i-1}}^{i-1}\\
\end{split}
\end{equation}
where $\cdot$ is the dot product.
We assume a subset of output features  $\mathcal{I}_{j}^{i-1}=\{o_{j}^{i-1},o_{x_1}^{i-1},o_{x_2}^{i-1},...,o_{x_m}^{i-1}\}$ are similar to each other  : 
\begin{equation}\label{features similar}
\begin{split}
o_{x0}^{i-1}\approx o_{x_1}^{i-1}\approx o_{x_2}^{i-1} \approx ... \approx o_{x_m}^{i-1},\\ s.t.\  0\leq m\leq n^{i-1}
\end{split}
\end{equation}
The set of filters corresponding to $\mathcal{I}_{j}^{i-1}$ is $\mathcal{C}_{j}^{i-1}=\{f_{j}^{i-1},f_{x_1}^{i-1},f_{x_2}^{i-1},...,f_{x_m}^{i-1}\}$
So,we can reformulate Eq.\,(\ref{output feature}) as:
\begin{equation}\label{output feature reformulate}
\begin{split}
o_j^{i}=  f_j^i \cdot (\mathcal{O}_{\mathcal{L}^{i-1}} - \mathcal{I}_{j}^{i-1} ) +  f_j^i \cdot \mathcal{I}_{j}^{i-1}
\end{split}
\end{equation}
and,
\begin{equation}\label{output feature simple}
\begin{split}
f_j^i  \cdot \mathcal{I}_{j}^{i-1} =( k_{j}^{i,j}+  k_{x1}^{i,j}+  k_{x2}^{i,j}+...+k_{x_m}^{i,j})  \\ \cdot \{o_{j}^{i-1},o_{x_1}^{i-1},o_{x_2}^{i-1},...,o_{x_m}^{i-1}\}
\end{split}
\end{equation}
Combining Eq.\,(\ref{features similar}) and Eq.\,(\ref{output feature simple}) we see that,
\begin{equation}\label{approx}
\begin{split}
f_j^i  \cdot \mathcal{I}_{j}^{i-1} \approx ( k_{j}^{i,j}+  k_{x1}^{i,j}+  k_{x2}^{i,j}+...+k_{x_m}^{i,j})*o_{j}^{i-1}
\end{split}
\end{equation}
Hence, we let 
\begin{equation}\label{adjusting}
\begin{split}
k_{j}^{i,j} = k_{j}^{i,j}+  k_{x1}^{i,j}+  k_{x2}^{i,j}+...+k_{x_m}^{i,j}
\end{split}
\end{equation}
we get:
\begin{equation}\label{output feature subset}
\begin{split}
f_j^i  \cdot \mathcal{I}_{j}^{i-1} \approx k_{j}^{i,j}*o_{j}^{i-1}
\end{split}
\end{equation}
So we only need to keep $o_{j}^{i-1}$ and prune the others. In filter pruning of the $(i-1)$-th layer, the filter $f_{j}^{i-1}$, corresponding to $o_{j}^{i-1}$, will be reserved, and the others will be pruned.
According to Eq.\,(\ref{adjusting}), we directly adjust the weight of filter $f_{j}^{i}$.
Eventually, we obtain a smaller computational model, as described in Eq.\,(\ref{output feature subset}).
Overall, if the elements in $\mathcal{I}_{j}^{i-1}$ are the same, then $f_{j}^{i-1}$, which generates $o_{j}^{i-1}$, can be regarded as a central filter that covers the function of other filters in $\mathcal{C}_{j}^{i-1}$, and without obvious loss of accuracy. 
%
%In other words, $f_{j}^{i-1}$ is approximately equal all other filters in $\mathcal{C}_{j}^{i-1}$.
%
Some of the previous works used redundancy to pruning, but they rely on an assumption: Redundancy is useless, which is a lack of rigorous mathematical proof.
We use the similarity of the output feature maps and directly adjust the corresponding filters. 
The additional loss is mainly derived from the similar evaluation error of the output feature maps.
%
%-------------------------------------------------------------------------
\subsection{Filter Selection}\label{Filter Selection}
One critical problem is how to determine central filters, which respectively replace a subset of filters while minimizing the loss as much as possible.
To address it, we explore the relationship between features.
For any feature map $o_{j}^{i}$ in the $i$-th layer, we set a threshold $\theta^i $ and calculate the similarity between $o_{j}^{i}$ and other feature maps. If it is less than the threshold  $\theta^i $, then the undirected edges are connected between them so that a similarity graph $D_j^i$ is obtained.
By corresponding the similarity graph $D_j^i$ on the feature map $o_{j}^{i}$ to the filter $f_{j}^{i}$, we get the relationship graph $G_j^i$ between the filters and the set of points is defined as $\mathcal{C}_j^i$. Furthermore, the size of set $\mathcal{C}_j^i$ ranges from 1 to $n^i$. 
So our method can be formulated as an optimization problem:
\begin{equation}\label{optimization}
\begin{split}
min(\sum\limits_{i=1}^{N}\sum\limits_{j=1}^{n^i}\delta_j^i \mathcal{J}(f_j^i,\mathcal{C}))
\end{split}
\end{equation}
where $\delta_j^i$ is an indicator which is $1$ if $f_j^i$ is a central filter, otherwise is $0$.
$\mathcal{J}$  measures the loss caused by using $f_j^i$ to represent $\mathcal{C}$. 
Besides, for a defined compression rate, there are a certain number of central filters corresponding to it. A central filter may only replace itself when its output feature map is not similar to the others. In contrast, it can also replace more. 
Thus Eq.\,(\ref{optimization}) is equal to finding an optimal set of central filters to minimize the loss.

We adopt centrality evaluation to solve this problem.
In a connected graph, the closeness centrality of a node is a measure of centrality in a network, calculated as the reciprocal of the sum of the length of the shortest paths between the node and all other nodes in the graph. Thus, the more central a node is, the closer it is to all other nodes and the value is greater.
Similarly, in the similarity graph of features, the greater a feature's closeness centrality is, the more similar it is to all other features. 
We define the closeness centrality of feature $o_{j}^i$ as:
\begin{equation}\label{closeness centrality}
\begin{split}
c_j^i=\frac{n}{\sum\limits_{o\in \mathcal{I}_j^i ,o\neq o_{j}^i}^{}\mathcal{S}(o_{j}^i,o)}
\end{split}
\end{equation}
where $n+1$ is the size of set $\mathcal{I}_j^i -o_{j}^i$. 
For filter selection strategy, we iteratively select the filter corresponding to the feature map with the highest closeness centrality as the central filter, and then prune the other filters that are similar to its feature map.
The number of central filters is determined by a defined compression rate.   
As discussed in Sec.\;\ref{CentralFilter}, the error of our method mainly comes from the similarity measurement error.
Hence, we preferentially select the filters with the highest closeness centrality as the central filter based on the greed principle. 
Besides, for filters that will be pruned, selecting a central filter with the most similar feature map is more in line with our proposed method.
Overall, we rank the filters according to their closeness centrality from highest to lowest and then prune them in order, so that we can obtain the approximate optimal solution of Eq.\,(\ref{optimization}).
As shown in Fig.\;\ref{central filter process}, filters A, B, and C are similar to each other. Assuming that the closeness centrality of A is greater than B and C, which means A is more similar to others averagely. Then, we adjust $A = A + B + C$ and prune B and C. 
To illustrate, we plot the comparison of the accuracy before and after the adjustment in Fig.\;\ref{filter adjust}.  
It can be observed that the accuracy of the adjusted model has improved to varying degrees, which is affected by whether the feature maps are sufficiently similar. 
The more similar the feature maps, the greater the degree of improvement.
%
%-------------------------------------------------------------------------
\subsection{Pruning Procedure}\label{Pruning_Procedure}
We summarize our pruning procedure as follows:
First, based on the stability of the similarity between features, we estimate the average similarity between each pair of output feature maps for each layer. Then, we get the similarity matrix $S^i$.
Second, according to the defined compression rate, we calculate the threshold $\theta^i $ of each layer. Then, we can get the relationship graph  $G^i$ through $\theta^i$ and $S^i$.
Third, we calculate the closeness centrality $\mathcal{C}^i$ of all nodes(filters) of $G^i$ and determine which filters should be pruned. For each filter $f_j^i$ that has not been pruned, we calculate its $\mathcal{C}_j^i$ according to the relationship in graph $G^i$.
Fourth, we adjust the corresponding weight according to $f_j^i$ and $\mathcal{C}_j^i$, as shown in Fig.\;\ref{central filter process} and Eq.\,(\ref{adjusting}).
Lastly, we iteratively prune similar filters and adjust the corresponding weight of filters of the next layer, and then through fine-tuning to recover the accuracy.
%
%-------------------------------------------------------------------------
\section{Experiments} \label{experiment}
\begin{figure}[!t]
\begin{center}
\hspace{-1.3em}
\includegraphics[height=0.65\linewidth]{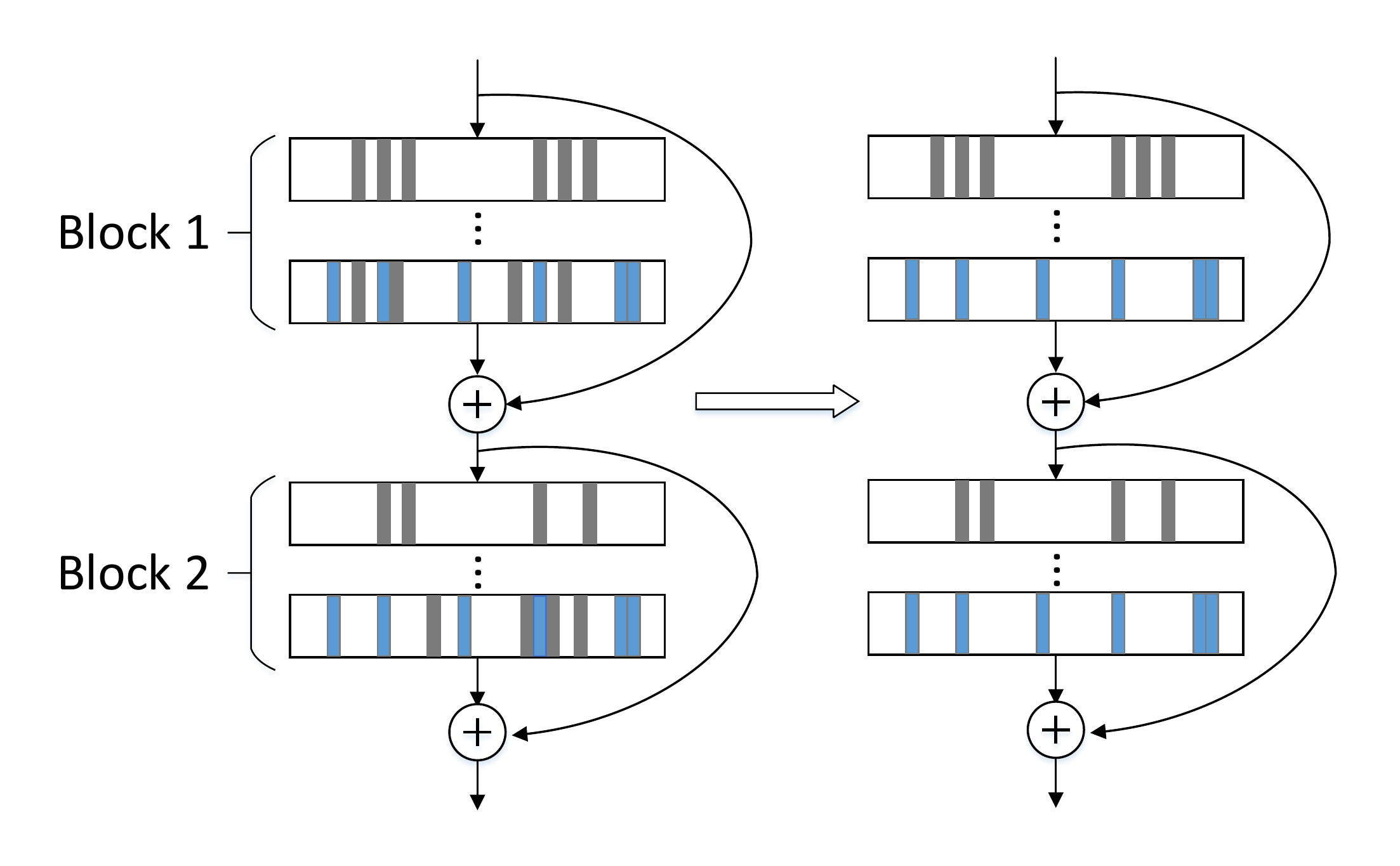}
\end{center}
\vspace{-1.2em}
\caption{\label{union_rank}
The pruning strategy of ResNet. Each large black rectangle represents a convolution layer, and each small rectangle inside represents a filter that will be pruned. The indexes of filters, which will be pruned, in the last convolutional layer of each block should be the same, because of the residual connection, and the blue rectangle denotes the intersection.
\vspace{-1.2em}
}
\end{figure}

To demonstrate the effectiveness of our proposed method in model compression, we conduct extensive experiments on CIFAR-10 \cite{Krizhevsky2009LearningML} and ImageNet \cite{Krizhevsky2012ImageNetCW}, using VGGNet \cite{Simonyan2015VeryDC}, GoogLeNet \cite{Szegedy2015GoingDW}, DenseNet \cite{Huang2017DenselyCC}, and ResNet \cite{He2016DeepRL}, for image classification.
We randomly sample 128 images (output feature maps) to estimate the average similarity of each layer of the different architectures on CIFAR-10. The difference is that we sample 16 images on ImageNet to save time. 
Furthermore, different architectures have different characteristics, such as ResNet with a residual block, GoogLeNet with an inception module, etc. So the pruning strategies are also different.
For all architectures, we sample output feature maps in ReLU \cite{Glorot2011DeepSR} layer before the next convolutional layer, prune filters in this layer, and adjust the corresponding weights in the next layer.
In particular, for ResNet, due to the shortcut connection, it is necessary to keep the two feature maps in the same dimension and the correspondence between the channels remains unchanged after pruning, as shown in Fig.\;\ref{union_rank}.
%
%-------------------------------------------------------------------------
\subsection{Experimental Settings}
\textbf{Configurations}.
In this paper, we conduct all pruning experiments on Pytorch 1.8 \cite{Paszke2017AutomaticDI} under Intel i7-8700K CPU @3.70GHz and NVIDIA GTX 1080Ti GPU, and use Stochastic Gradient Descent algorithm (SGD) for the optimization problem. 
For VGGNet-16, DenseNet-40, and ResNet-56, we set initial learning rate, batch size, momentum, and weight decay to 0.001, 128, 0.9 and 0.005, respectively.
We fine-tune the model for 30 epochs after pruning of each layer and set the learning rate to 0.0001 at epoch 5.
To improve the efficiency of pruning and save training time, we fine-tune GoogLeNet / ResNet-50 for 1-2 epochs after pruning of each inception/block to get the final network model.  
Then, the final model is fine-tuned for 30-40 epochs, in which the initial learning rate, batch size, momentum, and weight decay are set to 0.001, 64, 0.9, and 0.0005, respectively. We divide the learning rate by 10 at epochs 5,20.
For a fair comparison, we fix FLOPs and parameters as benchmarks and measure the Top-1 and Top-5 accuracy. Then, we compare them with the previous methods \cite{Li2017PruningFF,Huang2018DataDrivenSS,Lin2019TowardsOS,Lin2020HRankFP,He2017ChannelPF,Liu2017LearningEC,Lin2018AcceleratingCN,Luo2017ThiNetAF,Zhao2019VariationalCN}.
%
%-------------------------------------------------------------------------
\subsection{Results and Analysis}\label{results_and_analysis}

\subsubsection{Results on CIFAR-10}\label{results_on_cifar10}
We conduct experiments with several mainstream CNN models on CIFAR-10, including VGG-16, GoogLeNet, ResNet56, and DenseNet-40. The architecture of VGG-16 that we used is the same as \cite{Li2017PruningFF}. Moreover, We adjust the output of the original GoogLeNet matching the number of categories of CIFAR-10 \cite{Lin2019TowardsOS}.
%-------------------------------------------------------------------------

\textbf{VGG-16}.
The experimental results of VGG-16 are summarized in Tab.\;\ref{vggnet_cifar10}.
Compared with L1, CF significantly reduces the complexity of the model (60.4\% vs. 34.3\% for FLOPs and 83.6\% vs. 64.0\% for parameters) with just a 0.29\% loss in accuracy, whereas L1 suffers a 0.56\% loss in accuracy. 
CF outperforms HRank and Zhao \emph{et al}. in terms of acceleration rate (60.4\% vs. 53.5\% by HRank and 39.1\% by Zhao \emph{et al}.), while maintaining higher accuracy (93.67\% vs. 93.43\% by HRank and 93.18\% by Zhao \emph{et al}.).  
Compared with GAL-0.05 and SSS, our method achieves better accuracy (93.06\% vs. 92.03\% by GAL-0.05 and 93.02\% by SSS.) under higher parameters (85.4\% vs. 77.6\% by GAL-0.05 and 73.8\% by SSS.) and FLOPs (70.8\% vs. 39.6\% by GAL-0.05 and 41.7\% by SSS.) reduction.
Besides, compared with HRank, CF yields a better accuracy (92.49\% vs. 91.23\%) and a greater FLOPs reduction (79.0\% vs. 76.5\%) under the similar parameter reduction (89.9\% vs. 88.1\%).
This demonstrates that our method is more likely to yield better results when highly efficient models are required.  
\begin{table}[]
\scriptsize
\centering
\caption{Pruning results of VGGNet on CIFAR-10.}
\label{vggnet_cifar10}
\begin{tabular}{cccc}
\toprule
Model                                                 &Accuracy(\%)           &FLOPs(PR)         &Parameters(PR)\\
\midrule
VGGNet                                                &93.96             &313.73M(0.0\%)    &14.98M(0.0\%)  \\
L1 \cite{Li2017PruningFF}                               &93.40             &206.00M(34.3\%)    &5.40M(64.0\%) \\
Zhao \emph{et al.} \cite{Zhao2019VariationalCN}         &93.18             &190.00M(39.1\%)    &3.92M(73.3\%) \\
GAL-0.05 \cite{Lin2019TowardsOS}                        &92.03             &189.49M(39.6\%)    &3.36M(77.6\%)  \\
SSS \cite{Huang2018DataDrivenSS}                      &93.02             &183.13M(41.6\%)    &3.93M(73.8\%) \\
%
%GAL-0.1 \cite{lin2019towards}                         &90.73             &171.89M(45.2\%)    &2.67M(82.2\%)\\
%
HRank \cite{Lin2020HRankFP}                        &93.43             &145.61M(53.5\%)    &2.51M(82.9\%)   \\
\textbf{CF}(Ours)					        &93.67			&124.10M(60.4\%)   &2.45M(83.6\%)  \\
HRank  \cite{Lin2020HRankFP}                      &92.34             &108.61M(65.3\%)    &2.64M(82.1\%)   \\
\textbf{CF}(Ours) 					      &93.06			&91.54M(70.8\%)   &2.18M(85.4\%)  \\
HRank  \cite{Lin2020HRankFP}                     &91.23             &73.70M(76.5\%)     &1.78M(92.0\%)   \\
\textbf{CF}(Ours) 					      &92.49			&65.92M(79.0\%)   &1.51M(89.9\%)  \\
\bottomrule
\end{tabular}
\vspace{-1.0em}
\end{table}
%
%-------------------------------------------------------------------------

\textbf{GoogLeNet}.
The results on GoogLeNet are displayed in Tab.\;\ref{googlenet_cifar10}.
Our proposed method achieves a 63.2\% FLOPs reduction with only a small loss of 0.35\% in the top-1 accuracy for CIFAR-10, which demonstrates its ability to reduce network complexity.
Compared with GAL-0.05 and HRank, CF achieves higher FLOPs (63.2\% vs. 38.2\% by GAL-0.05 and 54.9\% by HRank) and parameters (55.6\% vs. 49.3\% by GAL-0.05 and 55.4\% by HRank) reduction, while maintaining higher accuracy (94.70\% vs. 93.93\% by GAL-0.05 and 94.53\% by HRank).
In general, CF is more efficient in accelerating neural networks. Hence, it can also be used to compress networks with inception modules. 
\begin{table}[]
\scriptsize
\centering
\caption{Pruning results of GoogLeNet on CIFAR-10.}
\label{googlenet_cifar10}
\begin{tabular}{cccc}
\toprule
Model                                                 &Accuracy(\%)           &FLOPs(PR)       &Parameters(PR)\\
\midrule
GoogLeNet                                              &95.05             &1.52B(0.0\%)    &6.15M(0.0\%)     \\
L1 \cite{Li2017PruningFF}                               &94.54             &1.02B(32.9\%)    &3.51M(42.9\%) \\
Random                                                 &94.54             &0.96B(36.8\%)    &3.58M(41.8\%)      \\
GAL-0.05 \cite{Lin2019TowardsOS}                         &93.93             &0.94B(38.2\%)    &3.12M(49.3\%)  \\
%
%GAL-ApoZ \cite{hu2016network}                         &92.11             &0.76B(50.0\%)    &2.85M(53.7\%)  \\
%
HRank \cite{Lin2020HRankFP}                           &94.53             &0.69B(54.9\%)    &2.74M(55.4\%)   \\
\textbf{CF}(Ours)                                     &94.70             &0.56B(63.2\%)    &2.73M(55.6\%)   \\
HRank \cite{Lin2020HRankFP}                          &94.07             &0.45B(70.4\%)    &1.86M(69.8\%)   \\
\textbf{CF}(Ours)                           	     &94.13             &0.37B(75.7\%)    &2.17M(64.7\%)  \\
\bottomrule
\end{tabular}
\vspace{-1.5em}
\end{table}
%
%-------------------------------------------------------------------------

\textbf{ResNet-56}.
Tab.\;\ref{resnet56_cifar10} shows the result for ResNet-56.
Compared with L1 and HRank, CF achieves higher FLOPs reduction (31.4\% vs. 27.6\% by L1 and 29.3\% by HRank) while increasing the accuracy by around 0.38\%.
When compared with GAL-0.6 and He \emph{et al}., our method has a better performance in all aspects, which achieves better accuracy with fewer parameters and FLOPs.
Furthermore, compared with GAL-0.8 and Zhao \emph{et al}., CF yields a better accuracy (92.19\% vs. 90.36\% by GAL-0.8 and 90.80\% by Zhao \emph{et al}.), though it reduces more FLOPs (68.1\% vs. 60.2\% by GAL-0.8 and 50.6\% by Zhao \emph{et al}.).
This indicates that our proposed method can effectively compress networks with residual blocks.
\begin{table}[]
\scriptsize
\centering
\caption{Pruning results of ResNet-56 on CIFAR-10.}
\label{resnet56_cifar10}
\begin{tabular}{cccc}
\toprule
Model                                                &Accuracy(\%)           &FLOPs(PR)       &Parameters(PR)\\
\midrule
ResNet-56                                              &93.26              &125.49M(0.0\%)     &0.85M(0.0\%)     \\
L1 \cite{Li2017PruningFF}                               &93.06              &90.90M(27.6\%)   &0.73M(14.1\%) \\
HRank  \cite{Lin2020HRankFP}            			&93.52  			&88.72M(29.3\%)  &0.71M(16.8\%) \\
%
%NISP \cite{yu2018nisp}                                 &93.01              &81.00M(35.5\%)     &0.49M(42.4\%)  \\
%
\textbf{CF}(Ours)							&93.64			&86.11M(31.4\%)		&0.53M(37.6\%) \\
GAL-0.6 \cite{Lin2019TowardsOS}                      &92.98              &78.30M(37.6\%)   &0.75M(11.8\%)  \\
\textbf{CF}(Ours)							&93.59			&75.7M(39.7\%)		&0.45M(47.1\%) \\
HRank  \cite{Lin2020HRankFP}           			&93.17              &62.72M(50.0\%)  	&0.49M(42.4\%)   \\
He \emph{et al}. \cite{He2017ChannelPF}             &90.80              &62.00M(50.6\%)     &-      \\
GAL-0.8  \cite{Lin2019TowardsOS}                   &90.36              &49.99M(60.2\%)   &0.29M(65.9\%)  \\
\textbf{CF}(Ours)						    &92.19			&40.0M(68.1\%)	&0.19M(77.6\%) \\
%
%HRank                                 			&90.72           	&32.52M(74.1\%)   &0.27M(68.1\%)  \\
%
\bottomrule
\end{tabular}
\vspace{-1.5em}
\end{table}
%
%-------------------------------------------------------------------------

\textbf{DenseNet-40}.
Tab.\;\ref{densenet_cifar10}. summarizes the results on DenseNet-40.
CF outperforms HRank and GAL-0.01. in terms of acceleration rate (41.4\% vs. 40.8\% by HRank and 35.3\% by GAL-0.01), while maintaining higher accuracy (94.33\% vs. 94.24\% by HRank and 94.29\% by GAL-0.01).
Besides, compared with GAL-0.05 and Zhao \emph{et al}., CF yields a better accuracy (93.71\% vs. 93.53\% by GAL-0.05 and 93.16\% by Zhao \emph{et al}.), though it reduces more FLOPs (61.3\% vs. 54.7\% by GAL-0.05 and 44.8\% by Zhao \emph{et al}.).
This indicates that CF can effectively compress the model with dense blocks.
\begin{table}[]
\scriptsize
\centering
\caption{Pruning results of DenseNet-40 on CIFAR-10.}
\label{densenet_cifar10}
\begin{tabular}{cccc}
\toprule
Model                                                 &Accuracy(\%)            &FLOPs(PR)       &Parameters(PR)\\
\midrule
DenseNet-40                                            &94.81              &282.00M(0.0\%)    &1.04M(0.0\%)     \\
Liu \emph{et al.}-40\% \cite{Liu2017LearningEC}        &94.81              &190.00M(32.8\%)    &0.66M(36.5\%)      \\
%
%Liu \emph{et al.}-70\% \cite{liu2017learning}          &94.35              &120.00M(57.6\%)    &0.35M(66.3\%) \\
%
GAL-0.01 \cite{Lin2019TowardsOS}                  &94.29              &182.92M(35.3\%) &0.67M(35.6\%)  \\
HRank  \cite{Lin2020HRankFP}                   &94.24                   &167.41M(40.8\%)  &0.66M(36.5\%) \\
\textbf{CF}(Ours)                                   &94.33            &165.38M(41.4\%)   &0.67M(35.6\%) \\
Zhao \emph{et al.} \cite{Zhao2019VariationalCN}       &93.16              &156.00M(44.8\%)&0.42M(59.7\%) \\
GAL-0.05 \cite{Lin2019TowardsOS}                &93.53              &128.11M(54.7\%) &0.45M(56.7\%)  \\
%
%\textbf{HRank}(Ours) &93.68 &110.15M(61.0\%) &0.48M(53.8\%) \\
%
\textbf{CF}(Ours)                                   &93.71              &109.40M(61.3\%)   &0.46M(55.8\%) \\
%
%GAL-0.1 \cite{lin2019towards}                          &91.90              &80.89M(71.4\%)  &0.26M(75.0\%)\\
%
\bottomrule
\end{tabular}
\vspace{-1.5em}
\end{table}
%
%-------------------------------------------------------------------------
\subsubsection{Results on ImageNet}\label{results_on_imagenet}
The experimental results of  ResNet-50 are summarized in Tab.\;\ref{resnet_imagenet}.  
Compared with GAL-0.5 and HRank, CF achieves higher FLOPs (47.9\% vs. 43.0\% by GAL-0.5 and 43.8\% by HRank) and parameters (36.9\% vs. 16.9\% by GAL-0.5 and 36.7\% by HRank) reduction, while maintaining better top-1 accuracy (75.08\% vs. 71.95\% by GAL-0.5 and 74.98\% by HRank).
In terms of FLOPs and parameter pruning, CF outperforms other proposed methods such as SSS-26, SSS-32, and He \emph{et al}. in all aspects, while it still yields better top-1 and top-5 accuracy.
With similar FLOPs reductions, CF achieves better performance than GAL-1 and HRank, while reducing more parameters.
Moreover, CF still performs well at high compression rates.
For example,
when compared with ThiNet-50, CF gains better top-1 (70.26\% vs. 68.42\%) and top-5 (89.82\% vs. 88.30\%) accuracies, though it reduces more FLOPs (76.2\% vs. 73.1\%).
Similar results can be found when compared with HRank and GAL-1-joint.
The results show that CF has excellent performance on the large-scale ImageNet dataset.
\begin{table}[]
\scriptsize
\centering
\setlength{\tabcolsep}{4.7pt}
\caption{Pruning results of ResNet-50 on ImageNet.}
\label{resnet_imagenet}
\begin{tabular}{ccccc}
\toprule
Model                                    &Top-1\%  &Top-5\%   &FLOPs(PR)     &Parameters(PR)\\
\midrule
ResNet-50 \cite{Luo2017ThiNetAF}         &76.15    &92.87     &4.09B(0.0\%)     &25.50M(0.0\%)     \\
SSS-32\cite{Huang2018DataDrivenSS} 	   &74.18   		&91.91		&2.82B(31.1\%)		&18.60M(27.1\%)\\
He et al.\cite{He2017ChannelPF}         &72.30  		&90.80		&2.73B(33.3\%)		&-\\
GAL-0.5\cite{Lin2019TowardsOS} 		   &71.95  		&90.94		&2.33B (43.0\%)		&21.20M(16.9\%)\\
SSS-26\cite{Huang2018DataDrivenSS} 	   &71.82   		&90.79		&2.33B(43.0\%)		&15.60M(38.8\%)\\
HRank\cite{Lin2020HRankFP} 		       &74.98   		&92.33		&2.30B(43.8\%)		&16.15M(36.7\%)\\
\textbf{CF}(Ours)                       &75.08         &92.30           &2.13B(47.9\%)       &16.08M(36.9\%)\\
GDP-0.6\cite{Lin2018AcceleratingCN} 	   &71.19  		&90.71		&1.88B(54.0\%)		&-\\
%
%SCOP\cite{Tang2020SCOPSC} 			   &75.26   		&92.53 		&1.85B(54.6\%) 		&12.29M (51.8\%)\\
%
GAL-0.5-joint\cite{Lin2019TowardsOS}   &71.80   		&90.82		&1.84B(55.0\%)		&19.31M(24.3\%)\\
GAL-1\cite{Lin2019TowardsOS} 		  &69.88  		&89.75		&1.58B (61.4\%)		&14.67M(42.5\%)\\
%
%GDP-0.5\cite{Lin2018AcceleratingCN}  &69.58  		&90.14		&1.57B(61.6\%)		&-\\
%
HRank\cite{Lin2020HRankFP} 		    &71.98  		   &91.01		     &1.55B (62.1\%)		&13.77M(46.0\%)\\
\textbf{CF}(Ours)                   &73.43            &91.57            &1.50B(63.3\%)      &13.73M(46.2\%)\\
GAL-1-joint\cite{Lin2019TowardsOS} 	  &69.31  		&89.12 		&1.11B(72.9\%)		&10.21M(60.0\%)\\
ThiNet-50\cite{Luo2017ThiNetAF} 	  &68.42  		&88.30		&1.10B(73.1\%)		&8.66M(66.0\%)\\
HRank\cite{Lin2020HRankFP} 		     &69.10   		  &89.58		&0.98B(76.2\%)		&8.27M(67.6\%)\\
\textbf{CF}(Ours)                   &70.26            &89.82          &0.98B(76.2\%)      &8.10M(68.2\%)\\
\bottomrule
\end{tabular}
\vspace{-1.2em}
\end{table}
%
%-------------------------------------------------------------------------
\subsection{Ablation Study}\label{ablation_study}
We performed a systematic ablation study on the effectiveness of the filter selection strategy and the adjusting step. Besides, we show the changes in the similarity between feature maps after pruning.
\subsubsection{Pruning without Adjusting}\label{Pruning without Adjusting}

%%%%%%%%%%%%%%%%%%%%%%%%%%%%%%%%%%%%%%%%%%%%%%%%%%%%%%%%%%%%%%%%%%
\begin{figure}[!t]
\begin{center}

\hspace{0.2em}
\centerline{
\includegraphics[width=0.7\linewidth]{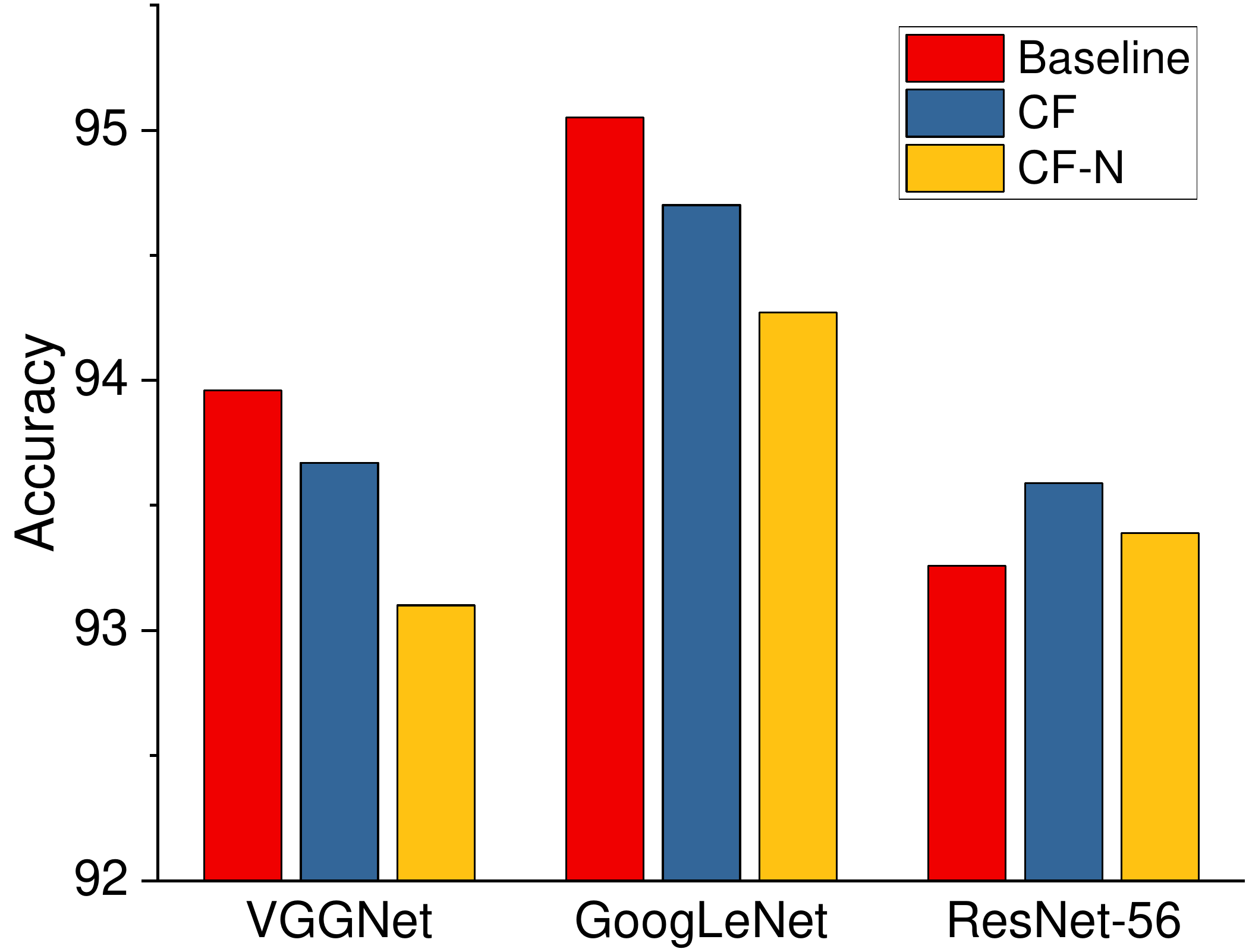}}%\hspace*{-0.18\linewidth}
\end{center}
\vspace{-0.8em}
\caption{\label{ablation-1}How adjusting weights in CF affects the top-1 accuracy. The CF-N denotes the CF without adjusting step.
}
\vspace{-0.8em}
\end{figure}
%%%%%%%%%%%%%%%%%%%%%%%%%%%%%%%%%%%%%%%%%%%%%%%%%%%%%%%%%%%%%%%%%%
%
We show that directly adjusting filters during fine-tuning yields a better performance on various benchmark models.
For brevity, we compare the results of the ResNet-56(93.59\% in Tab.\;\ref{resnet56_cifar10}), VGGNet(93.67\% in Tab.\;\ref{vggnet_cifar10}) and  GoogLeNet(94.70\% in Tab.\;\ref{googlenet_cifar10}) respectively.
For each model in Fig.\;\ref{ablation-1}, we observe that CF outperforms CF-N, which suggests that the adjusting step contributes to accuracy recovery. 
%
%-------------------------------------------------------------------------
\subsubsection{Variants of Filter Selection Strategy}\label{Variants of Filter Selection Strategy}
%%%%%%%%%%%%%%%%%%%%%%%%%%%%%%%%%%%%%%%%%%%%%%%%%%%%%%%%%%%%%%%%%%
\begin{figure}[!t]
\begin{center}
\hspace{0.2em}
\centerline{
\includegraphics[width=0.7\linewidth]{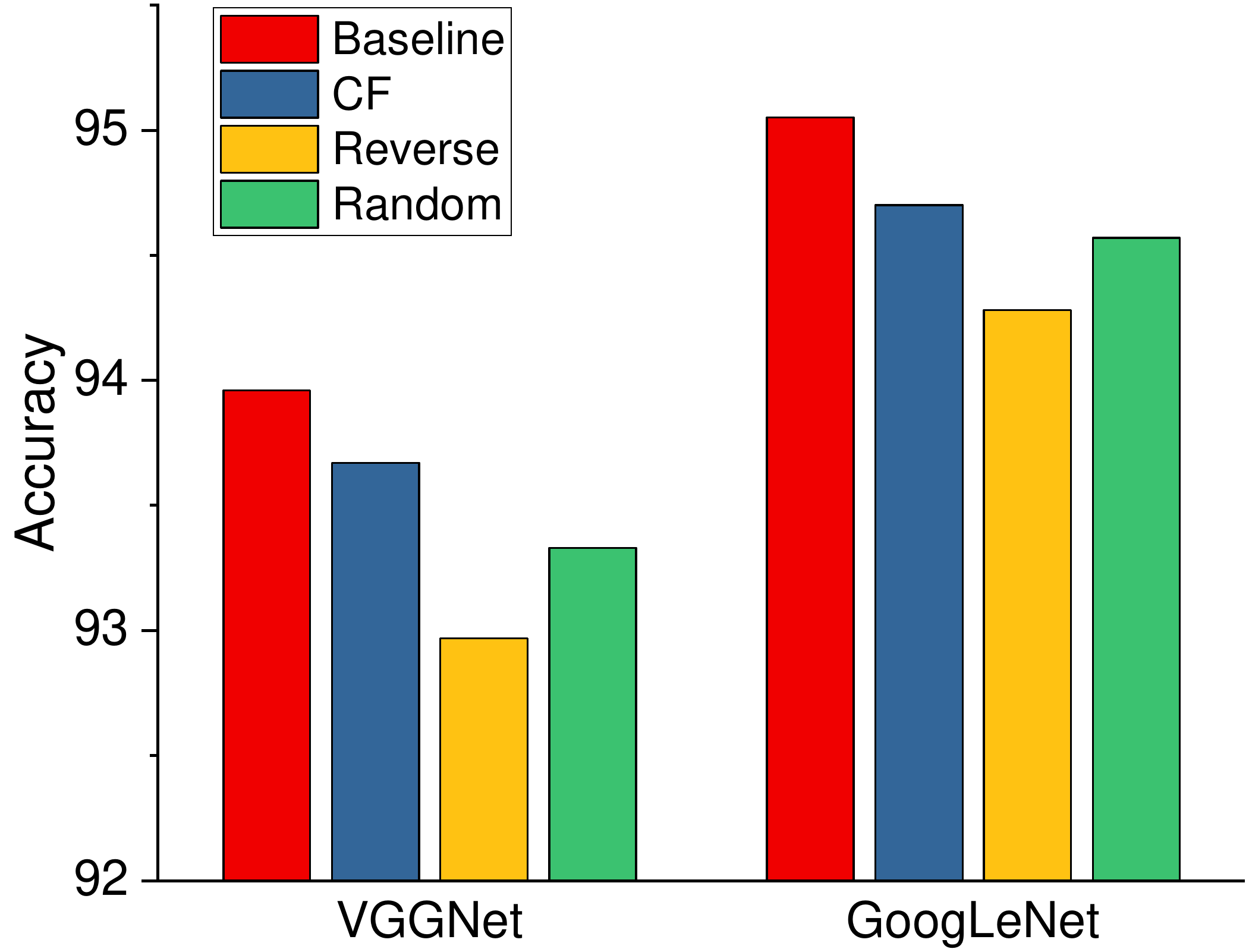}}%\hspace*{-0.18\linewidth}
\end{center}
\vspace{-0.8em}
\caption{\label{ablation-2}Top-1 accuracy for variants of filter selection strategy.
}
\vspace{-0.8em}
\end{figure}
%%%%%%%%%%%%%%%%%%%%%%%%%%%%%%%%%%%%%%%%%%%%%%%%%%%%%%%%%%%%%%%%%%
%
Fig.\;\ref{ablation-2} shows that preferentially preserving filters with higher closeness centrality of feature maps is superior.
We propose two variants: (1)Random: Filters are randomly pruned. (2)Reverse: Filters with higher closeness centrality are pruned first.
The pruning settings for VGGNet(93.67\%) / GoogLeNet(94.70\%) are the same as the before experiments in Tab.\;\ref{vggnet_cifar10},\ref{googlenet_cifar10}, respectively.
Among the variants, we observe that CF shows the best performance, while the Reverse CF has the worst performance, demonstrating that filters with higher closeness centrality are more suitable as central filter and can significantly reduce the additional errors caused by pruning.
%
%-------------------------------------------------------------------------
\subsubsection{Feature Similarity after Pruning}\label{Feature Similarity after Pruning}
\begin{figure}[!t]
\begin{center}

\hspace{0.2em}
\centerline{
\includegraphics[width=0.85\linewidth]{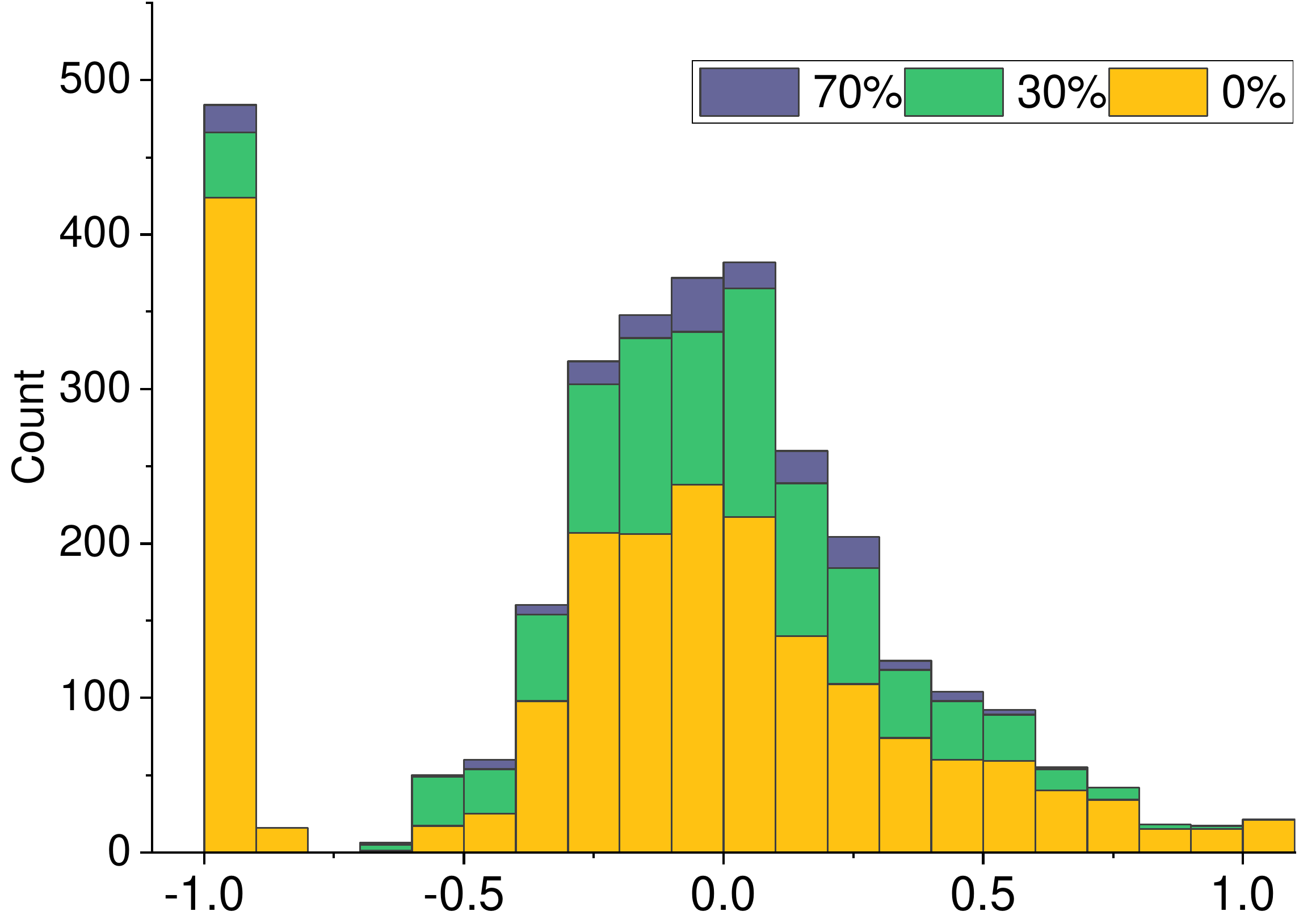}}
\end{center}
\vspace{-0.8em}
\caption{\label{ablation-3}Comparison of similarity between feature maps after pruning VGGNet at different compression rates. The x-axis is the similarity values, and the y-axis is the number of feature pairs.
}
\vspace{-1.2em}
\end{figure}

We show the comparison in the similarity between feature maps in the first layer of VGGNet at different compression rates, as shown in Fig.\;\ref{ablation-3}.
We can observe that the number of feature map pairs with similarity greater than 0.7 gradually decreases as the compression rate increases. When the compression rate is 70\%, the similarities between feature maps are all less than 0.6. 
This indicates that CF can effectively reduce the redundancy of the model.
%
%%%%%%%%%%%%%%%%%%%%%%%%%%%%%%%%%%%%%%%%%%%%%%%%%%%%%%%%%%%%%%%%%%
%%%%%%%%%%%%%%%%%%%%%%%%%%%%%%%%%%%%%%%%%%%%%%%%%%%%%%%%%%%%%%%%%%
%-------------------------------------------------------------------------
\section{Conclusions}\label{conclusion}
In this paper, we propose a novel filter pruning method called Central Filter (CF), which prunes filters by using the similarity information between feature maps.
Unlike the previous methods based on filter importance or redundancy, our approach mathematically proves that there are such filters that can replace other sets of filters after proper adjustment.
To support the proposed approach, we analyzed the average similarity of the output feature maps of the different models in CIFAR-10, which confirmed the feasibility of CF.
The advanced performance of CF in compressing networks is demonstrated by various experiments on benchmark network models.
In addition, we apply closeness centrality in graph theory to filter selection, proving its effectiveness through extensive ablation studies.
In future work, we will further study these existing or potential stable properties of feature maps.
%
%-------------------------------------------------------------------------
\section{Acknowledge}
This work was supported by National Natural Science Foundation of China (Grant Nos.61976246,U20A20227), Natural Science Foundation of Chongqing (Grant No.cstc2020jcyj-msxmX0385), National Key R\&D Program of China (Grant Nos.2018YFB1306600, 2018YFB1306604)
%-------------------------------------------------------------------------

%%%%%%%%% REFERENCES
{\small
\bibliographystyle{ieee_fullname}
\bibliography{egbib}
}

\end{document}